%% file: main-arxiv-2023.tex
\documentclass[11pt,a4paper]{article}
\usepackage{authblk}
\usepackage[acceptedWithA]{tacl2021v1}
\usepackage{times}
\usepackage{latexsym}

\usepackage{xcolor}

\newcommand\epochNumberStartingFromOne[1]{\number\numexpr1+#1\relax}

\newcommand\valscore[1]{#1}
\newcommand\mainresultstablePeerRead[1]{
\begin{tabular}{cccccc}
\hline
\multirow{2}{1cm}{Model} & \multicolumn{3}{c}{test scores} & \multicolumn{2}{c}{validation scores \& statistics}\\
 & Accuracy$\uparrow$ & ROC AUC$\uparrow$ & $F_{1}$-score$\uparrow$ & Accuracy$\uparrow$ & model epoch\\
 #1
\end{tabular}
}
\newcommand\mainresultstableACL[1]{
\begin{tabular}{cccccc}
\hline
\multirow{2}{1cm}{Model} & \multicolumn{3}{c}{test scores} & \multicolumn{2}{c}{validation scores \& statistics}\\
 & R2$\uparrow$ & MSE$\downarrow$ & MAE$\downarrow$ & R2$\uparrow$ & model epoch\\
 #1
\end{tabular}
}

\usepackage{microtype}

\usepackage{multirow}

\usepackage[T1]{fontenc}
\usepackage[acronym,shortcuts]{glossaries}
\setacronymstyle{long-short} 

\newsavebox{\largestimage}

\newcommand*{\acsu}[1]{\glsunset{#1}}  
\newcommand*{\acsun}[1]{\glsreset{#1}}

\usepackage{xcolor}
\usepackage{placeins}
\usepackage{subcaption}
\usepackage{adjustbox}
\usepackage{enumitem}
\usepackage{placeins}
\usepackage[noend]{algorithm2e}

\usepackage{colortbl}

\setlist[enumerate]{itemsep=0mm, topsep=3pt}
\setlist[itemize]{itemsep=0mm, topsep=3pt}
\newlist{questions}{enumerate}{2}
\setlist[questions,1]{label=RQ\arabic*.,ref=RQ\arabic*}
\setlist[questions,2]{label=(\alph*),ref=\thequestionsi(\alph*)}

\usepackage{here}     
\usepackage{tabu}
\usepackage{booktabs}
\newcommand{\midsepremove}{\aboverulesep = 0mm \belowrulesep = 0mm}
\midsepremove

\usepackage[title]{appendix}

\include{acronyms}

\newcommand\EXCLUDEEMNLPDRAFT[1]{}
\newcommand\EXCLUDE[1]{}

\title{MultiSChuBERT: Multi-Modal Scholarly Document Quality-Indicator Prediction Leveraging Pre-trained Models}
\title{MultiSChuBERT: Effective Multimodal Fusion for Scholarly Document Quality Prediction}

\makeatletter
\newcommand{\removelatexerror}{\let\@latex@error\@gobble}
\makeatother

\newcommand{\newchunk}[1]{\vspace*{1em}\noindent\textbf{#1}\noindent}

\newcommand{\addstretch}[1]{\addtolength{#1}{\fill}}

\usepackage[english]{babel}
\usepackage{blindtext}

\author[]{Gideon Maillette de Buy Wenniger$^{\textrm{ a,b}}$}
\author[]{Thomas van Dongen$^{\textrm{ a}}$}
\author[]{\textbf{Lambert Schomaker$^{\textrm{ a}}$}}
\affil[]{$^{\textrm{a }}$Bernoulli Institute for Mathematics,Computer Science and Artificial Intelligence \protect\\ \small University of Groningen, Groningen, The Netherlands \\
$^{\textrm{b} }$Faculty of Science, Open University, the Netherlands \\
 \texttt{gemdbw AT gmail.com}\\  \texttt{t.a.van.dongen AT student.rug.nl \ l.r.b.schomaker AT rug.nl}}

\date{}

\begin{document}

\maketitle
\begin{abstract}
\input{Abstract.tex}

\end{abstract}

\input{Introduction}
\input{RelatedWork}
\input{Datasets}
\input{CitationLabels}
\input{Methods}
\input{MainResultstablePeerRead}

\input{EvaluationMetrics}

\input{ResultsPyTorchReimplementation}

\FloatBarrier
\input{Conclusion}

\FloatBarrier
\clearpage

\bibliography{biblio}
\bibliographystyle{acl_natbib}

\clearpage
\begin{appendices}
\input{ChunkLengthAndOverlapAppendix.tex}
\input{MultiTaskLearningAppendix.tex}

\input{SChuBERTAppendix.tex}

\end{appendices}

\end{document}

%% file: acronyms.tex
\newacronym{LSTM}{LSTM}{long short-term memory}
\newacronym{BiLSTM}{BiLSTM}{bidirectional long short-term memory}
\newacronym{GRU}{GRU}{gated recurrent unit}
\newacronym{RNN}{RNN}{recurrent neural network}
\newacronym{NLP}{NLP}{natural language processing}
\newacronym{MT}{MT}{machine translation}
\newacronym{ML}{ML}{machine learning}
\newacronym{HAN}{HAN}{hierarchical attention network}
\newacronym[firstplural=hierarchical attention networks with structure tags]{HAN-ST}{HAN$_{\textrm{ST}}$}{hierarchical attention network with structure tags}
\newacronym{MSE}{MSE}{mean squared error}
\newacronym{MAE}{MAE}{mean absolute error}
\newacronym{SDQP}{SDQP}{scholarly document quality prediction}
\newacronym{BERT}{BERT}{Bidirectional Encoder Representations from
Transformers}
\newacronym{SciBERT}{SciBERT}{science BERT}
\newacronym{SChuBERT}{SChuBERT}{scientific chunking BERT}
\newacronym{MultiSChuBERT}{MultiSChuBERT}{multimodal scientific chunking BERT}

%% file: Abstract.tex
\acsu{BERT}
\acsu{SciBERT}
\acsu{SChuBERT}
\acsu{MultiSChuBERT}

Automatic assessment of the quality of scholarly documents is a difficult task with high potential impact. 
Multimodality, in particular the addition of visual information next to text, has been shown to improve the performance on \ac{SDQP} tasks. We propose the multimodal predictive model  \ac{MultiSChuBERT}.
It combines a textual model based on chunking full paper text and aggregating computed \ac{BERT} chunk-encodings (\ac{SChuBERT}), with a visual model based on Inception V3.
Our work contributes to the current state-of-the-art in \ac{SDQP} in three ways. First, we show that the method of combining visual and textual embeddings can substantially influence the results. Second, we demonstrate that gradual-unfreezing of the weights of the visual sub-model, reduces its tendency to ovefit the data, improving results. Third, we show the retained benefit of multimodality when replacing standard \ac{BERT}$_{\textrm{BASE}}$ embeddings with more recent state-of-the-art text embedding models. 

Using \ac{BERT}$_{\textrm{BASE}}$ embeddings, on the (log) number of citations prediction task with the ACL-BiblioMetry dataset, our \ac{MultiSChuBERT} (text+visual) model obtains an $R^{2}$ score of 0.454 compared to 0.432 for the \ac{SChuBERT} (text only) model. Similar improvements are obtained on the PeerRead accept/reject prediction task.
In our experiments using \ac{SciBERT}, scincl, SPECTER and SPECTER2.0 embeddings, we show that each of these tailored embeddings adds further improvements over the standard \ac{BERT}$_{\textrm{BASE}}$ embeddings, with the SPECTER2.0 embeddings performing best.


%% file: Introduction.tex
\acsun{SDQP}

\section{Introduction}

Is it possible to make automated quantitative predictions about the relative quality and impact of scholarly documents? 
The enormous growth in the number of publications has caused a growing demand from publishers and funding agencies, but to an extent also individual researchers for such \ac{SDQP}. Quantifying and predicting the \emph{real quality and impact} of scholarly documents, if such a thing exists, is as ambitious a goal as it is elusive. 
Even so, the famous aphorism ``\emph{All models are wrong but some are useful}'' \cite{BOX1979} seems to apply to SDQP as well.
This shows in the  growing interest in this research area, which in practice targets prediction of more attainable and measurable useful proxies of scholarly document quality, such as accept/reject labels or number of citations. 
Performance on \ac{SDQP} tasks has been steadily increasing. These performance gains are partly enabled by the emergence of powerful language models such as \ac{BERT} and its successors, and their more specialized variants.

SDQP models intuitively benefit from access to relevant and diverse forms of information pertaining to different quality aspects of the document content.
Multimodal predictive models that use both a visual and textual representation of the input \cite{shen2019joint} fit exactly this intuition. 
The rationale for the additional use of visual appearance in scholarly document evaluation is that visual features of carefully designed documents may contain subtle indicators of the maturity of the manuscript and, indirectly, the author's own trust in the quality of their findings.
Even so, combining textual and visual model components, a process called \emph{fusion}, in such a way that the two sources of information are complementary rather than mutually redundant is not straightforward.  \citet{shen2019joint} established the potential of multimodality for SDQP, but did not yet succeed to obtain consistent performance gains from it on the PeerRead \cite{PeerRead} accept/reject prediction task. This motivates our first research question:
\begin{questions}[itemindent=1em]
 \item How to best combine modalities in a multimodal model for \ac{SDQP} and train such a model in a way that achieves effective fusion of the multiple modalities? \label{RQ1}
\end{questions}

To achieve robust gains from multimodality requires two things: 1) It requires all the individual model subcomponents for the different modalities to be individually strong, 2) it requires a \emph{fusion} method that manages to combine and weigh the different subcomponents such that they can effectively complement each other in the final prediction.
This research provides scientific contributions to both required aspects. Contributing to the first requirement, textual, we apply powerful pre-trained Transformer models \cite{van_Dongen_2020} which facilitate full-text encoding, and show its effectiveness on the \ac{SDQP} task. 
Contributing to the second requirement, first, we show the importance of balancing different sub-models with large differences in the number of trainable parameters during training. In particular, we improve fine-tuning of pre-trained (visual) INCEPTION sub-models by regularizing it with gradual unfreezing.
Second, we show the importance of the choice of fusion technique when combining embeddings in a multi-modal setting.

Our multimodal prediction model MultiSChuBERT improves upon the current state-of-the-art on two sub-tasks of \ac{SDQP}, accept/reject prediction and number of citations prediction, including the multimodal work by 
\citet{shen2019joint}. The MultiSChuBERT model aggregates \ac{BERT}$_{\textrm{BASE}}$ embeddings for text chunks to represent scientific paper text,  and combines this with a representation of the paper pages organized in a (visual) image grid. 

Noting the reliance on \ac{BERT}$_{\textrm{BASE}}$ embeddings, 
an important remaining research question is: 
\begin{questions}[itemindent=1em]
\setcounter{questionsi}{1}
\item Does multimodality still benefit the prediction when applying powerful pre-trained language models specialized for the science domain? \label{RQ2}
\end{questions}
To answer this question, the last part of this work looks into the effect of (science) domain-specialized embedding models, as a replacement for \ac{BERT}$_{\textrm{BASE}}$. Looking at four domain-specialized embedding models, we find them all to further improve the results, but note that a part of these improvements may be due to label leakage.
However, for the best performing domain-specialized embedding model,  SPECTER2.0 \citep{cohan2020specter}, which precisely specifies the data used for training the model, we are able to run controlled experiments that avoid label-leakage. The results of these experiments show first that domain-specialized embeddings (still) substantially improve the results, when avoiding label leakage. Second, improvements of the multimodal model over the textual baseline add on to the (additional) improvements of the better textual embeddings, showing the comprehensive benefit of multimodality for \ac{SDQP}. \\

The rest of the paper is structured as follows. First, we provide an overview of related work. Next, we provide an overview of the used datasets in Section \ref{Section:Datasets} and a description of the number of citation labels in Section \ref{Section:Citation_Labels}. This is followed by an overview of the different models applied to the \ac{SDQP} tasks in Sections \ref{Section:Textual_Methods}--\ref{Section:Multi-task-Learning}, as well as hyperparameters and training techniques in Section \ref{Section:Hyper-parameters_and_Training_Techniques}. Next, the experiments are described in Section \ref{Section:Experiments}, followed by a description of the used evaluation metrics in Section \ref{Section:Evaluation_Metrics}. Finally, we describe the results and analysis of the experiments in Section \ref{Section:Results} followed by conclusions.

%% file: RelatedWork.tex
\acsun{SDQP}
\section{Related Work}

\Ac{SDQP} is a relatively new though fast growing field, 
\citet{sdpReview2021} provides a recent review.
This work also belongs to the broader field of multi-modal classification, of which a recent review is provided by \cite{MultimodalClassificationReview}. Closely related, though not exactly \ac{SDQP} are other forms of (multimodal) document quality prediction as well as what we will refer to as \emph{unrestricted} number of citations prediction which incorporates information such as author names, h-indexes etc, which we consider \emph{implicit labels} and therefore prohibit. While somewhat different in aim, certain recent work in the latter direction is technically advanced and very relevant to our research.
What follows is a discussion of selected recent works in the literature we found to be most relevant to the current work.

\subsection{Multimodal Document Quality Prediction}
An initial joint textual+visual modal for Wikipedia page quality prediction and the accept/reject prediction task of \ac{SDQP} was proposed by \citet{shen2019joint}. Their model encodes a (truncated) document text by representing sentences with average-word embedding and combining those with a BiLSTM. For the visual encoding, an Inception network applied to a fixed-size grid of pages is used.   
\citet{shenEtAl2020} extend upon this work, combining multi-modal document quality prediction with multi-task learning as well as the addition of hand-crafted features, showing improvement from both approaches.
Continuing in this direction, while focusing only on the Wikipedia quality prediction task, \citet{guda-etal-2020-nwqm} propose another improved multimodal approach for Wikipedia page quality prediction. Their model includes an INCEPTION-based visual encoding, following \cite{shen2019joint} and a textual encoding of the page sections similar to \citep{van_Dongen_2020}. In addition, it includes a representation of the information from the Talk page associated with each article, providing insight into the editorial state of the page.
They use the concatenation methods proposed by \citet{reimers2019sentencebert}, to combine the representations of the different information sources.

\citet{SelfMediaQualityPrediction} explore the related task of online self-media article quality prediction.
Their model combines three sub-models, encoding: 1) document text (BERT-based), 2) hand-crafted writing characteristics features, and 3) layout organization information obtained from page parsing. Their experiments show that the three sub-networks are complementary. 

 \citet{MultimodalGraphConvnetsHighQualityRecognition} perform the binary classification of social media articles into high/low-quality content. For this purpose, they use a graph convolutional network (GCN) that uses both the words from the article text and  the names of visual objects detected in article images. The GCN graph structure is heuristically determined based on point-wise mutual information determined by local word co-occurrence information.
 The shallow visual information only adds a small improvement over the text-only model in this approach. 

Compared with these earlier efforts, to the best of our knowledge, our work is the first to apply the multimodal Chunking-BERT textual + INCEPTION-based visual encoding architecture to \ac{SDQP} and show its merit for this task. In addition, the use of gradual unfreezing used in our work is shown to further boost performance. Finally, our work applies text embeddings specialized for the science domain as part of the multimodal prediction model. Our findings show these specialized embeddings to further improve performance in a way that is complementary to the gains obtained from the addition of the visual information.

\subsection{Number of Citations Prediction}
The task of number of citations prediction using unrestricted available information, including reviews, abstract text and other hand-crafted features such as authors' h-index is addressed by 
\citet{li-etal-2019-neural, CapsuleNetworksCitationPrediction}. 
\citet{li-etal-2019-neural} use a neural architecture that first combines the information from the abstract and each review (\emph{Abstract-Review match}) and then further combines the information of the thus re-weighed reviews (\emph{Cross-Review Match}). Technically, the network uses attention and average pooling, while creating initial word embeddings with word2vec and combining them to initial sentence embeddings using convolutional methods. Notably, their network performs  \emph{early fusion} by first producing abstract-aware review sentence representations and then combining those into full-review representations.
The more recent work of \citet{CapsuleNetworksCitationPrediction} is similar in its goal of combining abstract and review information effectively, but for this relies on capsule networks \cite{Hinton2011Transforming, DynamicRoutingBetweenCapsules} rather than attention. Both works combine the neural encoding with information from hand-crafted features as a final step, to further increase performance.

The number of citations prediction task as approached in these works differs from ours, as we prohibit \emph{implicit labels} such as review text or author information. Such information is not available in our \ac{SDQP} setting, where the goal is to predict quality indicators based on the document itself without the aid of implicit labels, akin to blind review. 

\subsection{Large Language Models for the Scientific Domain}

A number of domain-specialized large language models for the scientific domain have recently been proposed.
SciBERT \cite{Beltagy2019SciBERT} is a variant of \ac{BERT} fully trained on 1.14M full-text scientific papers from the computer science and biomedical domains. It includes a domain-specialized WordPiece vocabulary.  The model is shown to outperform \ac{BERT}$_{\textrm{BASE}}$ on several tasks, with the best results obtained when the model is fine-tuned for the specific task. SPECTER \cite{cohan2020specter} is another transformer model specifically created for the scientific domain. It is initialized using SciBERT, but is further trained using a citation-based pre-training objective. This training objective uses a triplet loss consisting of a query paper (the anchor), a positive paper (a paper that the query paper cites), and a negative paper (a paper that the query paper does not cite). Using this objective, SPECTER can learn document relatedness based on citations, which is a strong and important signal in the scientific domain. This model is shown to outperform SciBERT by a large margin on a number of tasks. SciNCL \cite{SCINCL} builds on top of the idea of SPECTER, but modifies the citation-based pre-training objective. Instead of relying on just one citation from the query to a positive paper, SciNCL uses citation embeddings, which use the full citation graph of the documents. By sampling positive and negative papers from the citation embedding graph space, they ensure that the positive and negative samples do not collide, but are also hard to learn. The model is shown to outperform SPECTER by a small margin. Recently, SPECTER 2.0 was published, based on \cite{cohan2020specter} and \cite{Singh2022SciRepEvalAM}. While it uses the same pre-training objective and base model as SPECTER, the new training data leads to a large improvement on the SciRepEval benchmark.

%% file: Datasets.tex
\section{Datasets}
\label{Section:Datasets}
A general overview of the datasets used in our experiments, including the number of documents and label types,  is given in Table \ref{datasets:sizes}. 
The datasets each have both a PDF (for the visual part) and parsed full-text (for the textual part) available.
For the datasets which contain accept/reject labels, the class distribution is listed in Table \ref{datasets:accept-reject}.

\begin{table*}[t]
\centering
\begin{subfigure}[b]{0.46\textwidth}

\scalebox{0.7}{
\begin{tabular}{lll}
\toprule
\textbf{Dataset}         & \begin{tabular}[c]{@{}l@{}}\textbf{\#Documents}\\ \textbf{(train + validation + test)}\end{tabular}            & \textbf{Labels}                   \\ 
\midrule
\rowcolor[HTML]{d1d1d1} 
AI              & 4092 (3682 + 205 + 205)                                
& Accept/reject            \\

CL              & 2638 (2374 + 132 + 132)                                                         & Accept/reject            \\

\rowcolor[HTML]{d1d1d1} 
LG              & 5048 (4543 + 252 + 253)                                                        & Accept/reject            \\

 \multirow{2}{1.5cm}{ACL- BiblioMetry} & \multirow{2}{*}{30950 (27853 + 1548 + 1549)}                                                    & \multirow{2}{*}{Citations}                \\
& & \\
\bottomrule
\end{tabular}}
\caption{Data sizes and label types}
\label{datasets:sizes}
\end{subfigure}
\begin{subfigure}[b]{0.49\textwidth}

\scalebox{0.7}{
\centering
\begin{tabular}{llll}

\textbf{Dataset}         & \begin{tabular}[c]{@{}l@{}}\textbf{Train} \\ \textbf{Accept : Reject}\end{tabular} & \begin{tabular}[c]{@{}l@{}}\textbf{Validation} \\ \textbf{Accept : Reject}\end{tabular}               & \begin{tabular}[c]{@{}l@{}}\textbf{Test} \\ \textbf{Accept : Reject}\end{tabular}                   \\ 
\midrule
\rowcolor[HTML]{d1d1d1} 
AI              & 10.5\% : 89.5\%                                 & 8.3\% : 91.7\%   & 7.8\% : 92.2\%              \\

CL              & 24.3\% : 75.7\%                                                        & 22.0\% : 78.0\% & 31.1\% : 68.9\%            \\

\rowcolor[HTML]{d1d1d1} 
LG              & 36.4\% : 63.6\%                                                        & 36.5\% : 63.5\% & 32.0\% : 68.0\%            \\

\bottomrule
\end{tabular}
}
\vspace{0.3cm}
\caption{PeerRead accept/reject distribution}
\label{datasets:accept-reject}
\end{subfigure}
\caption{Statistics of the used datasets. }
\label{datasets-classbalance}
\end{table*}

\subsection{PeerRead Dataset}
\label{PeerRead Dataset}
The PeerRead dataset is a dataset released in 2018 containing drafts, reviews and accept/reject decisions for 14784 papers. The dataset is divided into different subsets, of which the arXiv subset, which contains 11778 papers, is the largest. The arXiv dataset is further divided into three domains, namely Artificial Intelligence (cs.AI), Computation and Language (cs.CL), and Machine Learning (cs.LG). Hereafter, these will be referred to as AI, CL, and LG. As can be seen in table \ref{datasets-classbalance}, the PeerRead datasets are quite unbalanced. The AI dataset is the most unbalanced dataset, containing almost nine times as many rejected as accepted documents.

\subsection{ACL-BiblioMetry Dataset}
\label{ACL-BiblioMetry Dataset}
For number of citations prediction, we use the ACL-BiblioMetry dataset \cite{van_Dongen_2020}. 
The dataset contains 30950 full-text parsed papers split into 90\% training data and 5\% test and validation data. This dataset was created specifically for the citation prediction task. The dataset contains documents from two domains: Computational Linguistics and Natural Language Processing. Since the number of citations prediction task is difficult, it is beneficial to have a dataset that contains relatively similar documents. 

%% file: CitationLabels.tex
\section{Citation Labels}
\label{Section:Citation_Labels}
As explained in \cite{zipf_law}, the number of citations of scholarly documents follows Zipf-Mandelbrot's law. This means that citations over a set of documents increase according to the power law, where the most cited paper has twice as many citations as the second-most cited paper, the second-most cited paper twice as many citations as the third-most cited paper, etc. For this reason, the same methodology as proposed in \cite{structure_tags} is applied, where the function shown in equation \ref{citation_function} is applied to the number of citations for each document.

\begin{equation}
    \textrm{citation-score} = log_e(n + 1)
    \label{citation_function}
\end{equation}

\noindent Here, $n$ is the number of citations. One is added to $n$ to make sure that the function is well-defined for papers that do not have any citations. By taking the natural log of the citations, the effects of Zipf-Mandelbrot's law are mitigated. 

%% file: Methods.tex
\section{Textual Methods}
\label{Section:Textual_Methods}
The various methods used to predict the quality of scientific documents: 1) textual 2) visual and 3) joint (textual+visual);  are described in this and this and the following two sections.

\subsection{Textual Baseline Models}
\label{Textual Baseline Modles}
Multiple baseline prediction models that are used to compare the results of the proposed models with, are listed below. 

\begin{itemize}

  \item \Ac{BiLSTM}: the \ac{BiLSTM} model, as proposed in \cite{shen2019joint}, is a model which builds sentence representations by average pooling over the embedded words in a single sentence. The sentence representation is then used as input for a bidirectional LSTM, after which average pooling is applied a second time. 
  
  \item \Ac{HAN}: the \ac{HAN} model, as proposed in \cite{yang-etal-2016-hierarchical}. This model is used as a baseline for the citation prediction task.
\ac{HAN} starts from sentence-segmented text. It then first produces GloVe-based word embeddings for each sentence, and next uses a \ac{BiLSTM} with attention to produce sentence-embeddings for each sentence. A second document-level \ac{BiLSTM} with attention then combines these sentence embeddings into a document embedding. The document embedding is then further processed by a linear and softmax layer, producing class predictions.
  
  \item \Ac{HAN-ST}: the \ac{HAN} structure-tag model as proposed in \cite{structure_tags}. The model is similar to the \ac{HAN} model but uses structure tags that mark the role of sentences. The structure tags consist of TITLE, ABSTRACT, and BODY\_TEXT. Due to the extra implementation time, this model is not tested on the ACL dataset. However, since it outperforms \ac{HAN} on the accept/reject prediction task, it is used as a baseline rather than the standard \ac{HAN}.
  
\item Majority Training Label Prediction (Maj Training Label) and Average Training Label Prediction (Avg Training Label): baseline methods that predict always the majority class or average training label.
When looking at the results of predictive models on a certain \ac{SDQP} task, it is important to quantify how difficult that task really is. 
Results for these baseline prediction methods give a clear insight into the absolute and relative difficulty of the prediction tasks for certain datasets, given the available training data. Importantly, the results for these baselines also reflect how much the training and test data  are aligned, meaning they  have similar label distributions. This is important since higher similarity implies higher prior expected test performance for any model trained using the respective training data. 
\end{itemize}

\subsection{SChuBERT}
\label{SChuBERT}
The main model used for the textual part of this work is called \ac{SChuBERT} (Scholarly Chunking \ac{BERT}) \cite{van_Dongen_2020}. It consists of two main parts: a pre-trained \ac{BERT}$_{\textrm{BASE}}$ model and a deep learning model, which are both described in more detail below.\footnote{In the context of this paper, \ac{BERT}$_{\textrm{BASE}}$ refers to the uncased \ac{BERT}$_{\textrm{BASE}}$ model, in particular, the implementation available from \url{https://huggingface.co/bert-base-uncased}.} The pre-trained \ac{BERT} model is used to extract embeddings, while the deep learning model is used to learn from the embeddings and predict an output. \ac{BERT}$_{\textrm{BASE}}$ is pre-trained on BookCorpus \cite{zhu2015aligning} and English Wikipedia.

\begin{figure*}[t]
    \centering
    \scalebox{0.38}{
    \includegraphics{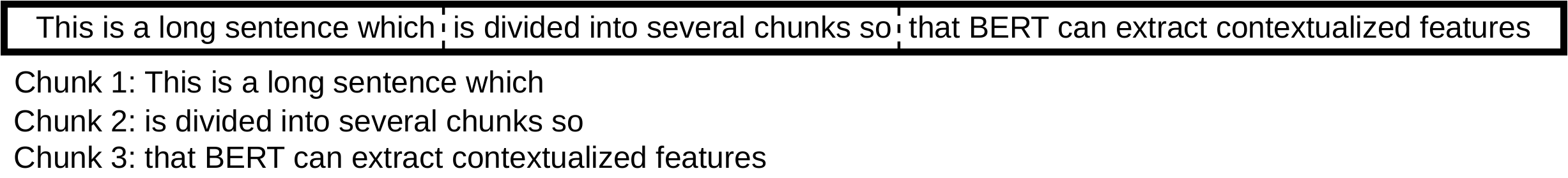}
    }
    \caption{Example of the chunking method. In this example, a sequence length of 6 used with no overlap.
    }
    \label{fig:chunking}
\end{figure*}

Transformers have a time complexity that is quadratic with respect to the input length \cite{vaswani2017attention}, therefore \ac{BERT} has a maximum input length of 512 tokens. Since scholarly documents are generally very long, it is impossible to use full documents as input for \ac{BERT}. For this reason, a method similar to the one proposed in \cite{m2019bert} is used. Documents are split into chunks of 512 tokens, optionally with an overlap of 50 tokens to preserve some context between the chunks. An example is shown in Figure \ref{fig:chunking}. These chunks are then used as input for the \ac{BERT}$_{\textrm{BASE}}$ model.

A pre-trained \ac{BERT}$_{\textrm{BASE}}$ model is used to generate embeddings for the chunks. Since we are interested in the hidden states output, a model is used without a task-specific dense layer. 

We use the \ac{BERT} embedding from the last layer, as shown by \citet{tune_or_no_tune} to be the most suitable both in the case of feature extraction (this work) as well as in the case of fine-tuning. \ac{BERT} outputs a length 768 vector for each token in the input (depending on which \ac{BERT} model is used). Therefore, mean-pooling over these token encodings is used to combine them into a single encoding, following \cite{reimers2019sentencebert} where mean-pooling is shown to outperform max-pooling. 
After the embeddings for all chunks of a single document are generated, they are merged into a single vector containing all chunks. The thus produced embeddings are generated once and then saved. This has some major computational advantages over approaches that fine-tune the entire \ac{BERT}-layer of the network.
Such approaches require re-computing example \ac{BERT} encodings for each experiment and each training epoch of that experiment. Compared to these approaches, our approach achieves major savings both in terms of computation and GPU memory usage.

To avoid overfitting, a simple architecture is used to further train with the generated document embeddings. Input vectors are of variable lengths since the number of chunks varies per document; this is accommodated using zero-padding and masking. Since the generated document embeddings have a sequential relation, a Gated Recurrent Unit (GRU) layer is used to learn from the document embeddings. The input is then passed through a dropout layer and finally a fully connected layer with a linear or softmax activation, depending on which task the model is trained on. 

In addition to the default BERT$_{\textrm{BASE}}$ embedding model, we also experiment with several more recent  embedding models that are more specifically trained for scientific documents. More details are given in the Experiment section (\ref{subsection:domain-specialized-bert-embedding}).

\section{Visual Methods}
\label{Visual Methods}
In this section, we describe the various image-based classification models and baselines as well as the used procedure for creating the visual datasets from raw PDF files. 

\subsection{Building the dataset}
\begin{figure*}[t]
   \begin{subfigure}[b]{0.40\textwidth}

    \centering
    \scalebox{0.27}{
    \includegraphics{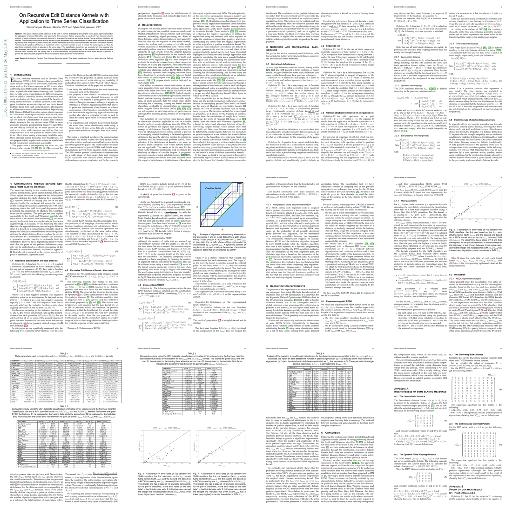}
    }
    \caption{Document grid -- overview.}
    \end{subfigure}
    \begin{subfigure}[b]{0.46\textwidth}
    \scalebox{2.5}{
       \includegraphics{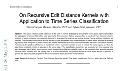}
    }
    \vspace{0.1cm}
    \caption{Top-left part of the document grid, containing title and abstract.}
    \end{subfigure}
    \caption{Example of created document grid. The grid contains 12 pages and is of size 512x512.}
    \label{fig:grid-example}
\end{figure*}

All of the datasets used in this work consist of raw PDF files. Since these are not suitable as inputs for an image classification network, they are first converted to 1700x2200 PIL \cite{clark2015pillow} image objects using pdf2image \cite{pdf2image}. Each image is first scaled down to 275x425. A grid is then created as shown in Figure \ref{fig:grid-example}. For the PeerRead dataset, the grid is 3x4, or 12 pages, for the ACL dataset it is 3x3, or 9 pages. These numbers, different for each dataset, are based on the average pages for each dataset. They are used to minimize truncation and padding while still using a fixed-size image as input for the model. 

Finally, the grids are resized to a single 512x512 image using bi-cubic interpolation.

\subsection{Visual Baseline Models}
\label{Visual Baseline Models}
As a baseline to compare the visual results against, the following baseline models are used:

\begin{itemize}
  \item{CNN: a simple CNN model. Since there are not many existing baselines apart from the INCEPTION baseline as described below, a simple network is tested which consists of a single convolutional layer followed by a global average pooling layer and one fully connected layer with either a linear or softmax activation. This is mainly included to show the improvement of the Inception architecture over a more traditional convolutional model.
  }
  \item{INCEPTION: the INCEPTION model as proposed in \cite{shen2019joint}. Architecturally, this model is identical to the proposed INCEPTION$_\text{GU}$ model, shown in Figure \ref{fig:visual-model}. 
  }

\end{itemize}

\subsection{INCEPTION$_\text{GU}$}
\label{INCEPTIONGU}
\begin{figure*}[ht]
    \begin{subfigure}[b]{0.55\textwidth}
    \scalebox{0.5}{
    \includegraphics{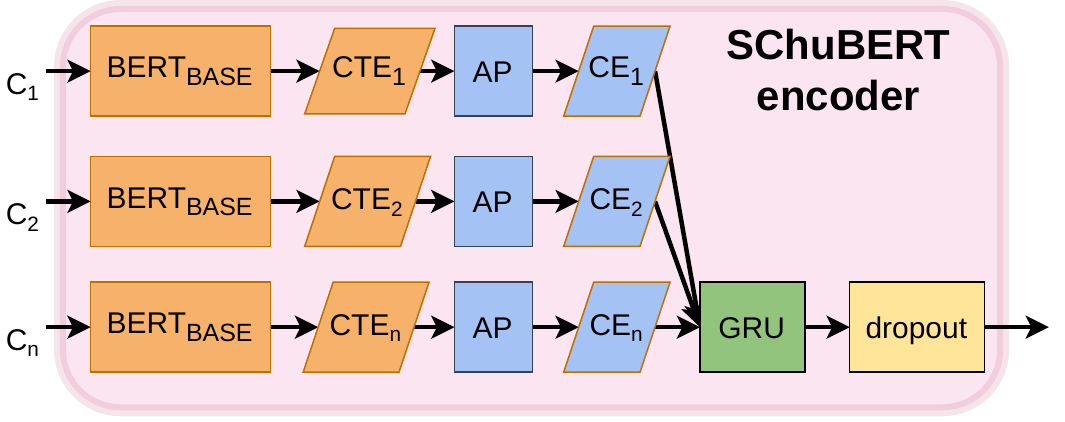}
    }
    \caption{SChuBERT encoder.}
    \end{subfigure}
    \begin{subfigure}[b]{0.39\textwidth}
    \scalebox{0.5}{
    \includegraphics{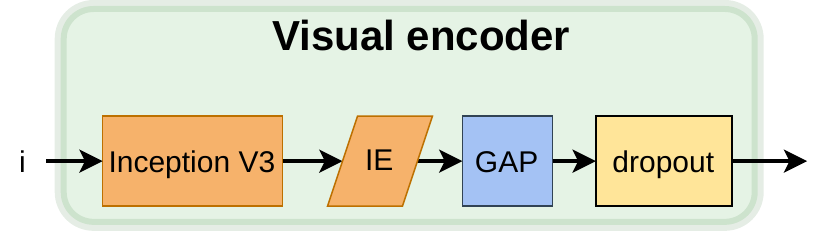}
    }
    \caption{Visual encoder.}
    \end{subfigure}

    \begin{subfigure}[b]{0.49\textwidth}
    \scalebox{0.5}{
    \includegraphics{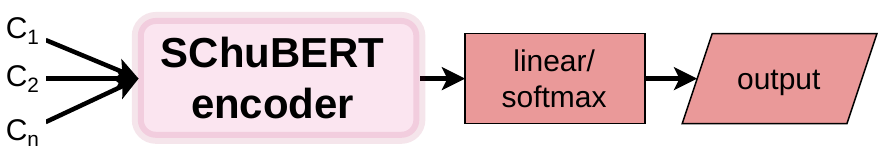}
    }
    \caption{The SChuBERT model \cite{van_Dongen_2020}.}
    \end{subfigure}
    \begin{subfigure}[b]{0.49\textwidth}
    \scalebox{0.5}{
    \includegraphics{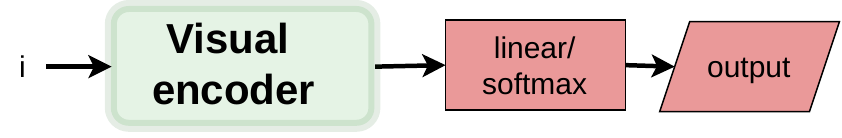}
    }
    \caption{The INCEPTION$_\text{GU}$ model proposed in this work, based of the INCEPTION model from \cite{shen2019joint}.}
    \end{subfigure}
    \begin{subfigure}[b]{0.69\textwidth}
    \centering
    \scalebox{0.5}{
    \includegraphics{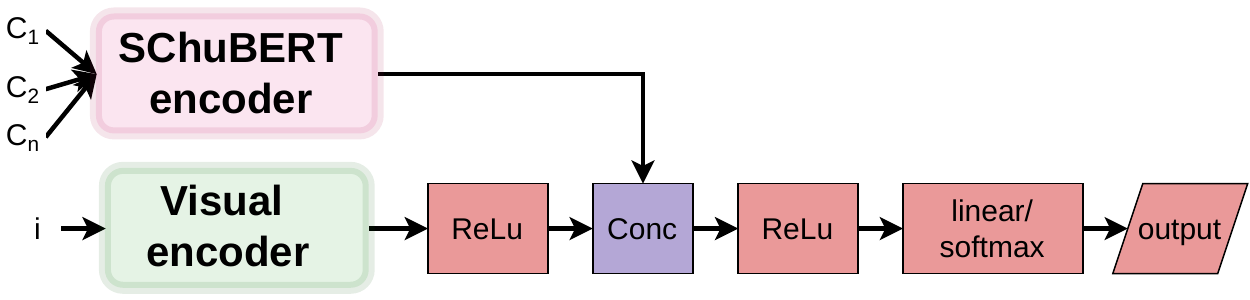}
    }
     \caption{The MultiSCHuBERT model proposed in this work.}
    \label{fig:schubert-joint-model}
    \end{subfigure}
    \begin{subfigure}[b]{0.3\textwidth}
    \centering
    \scalebox{0.5}{
    \includegraphics{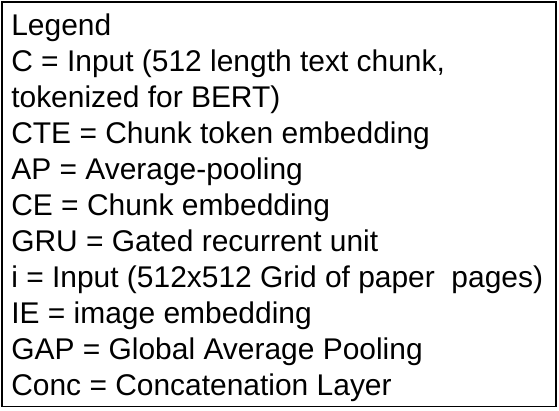}
    }
     \caption{Legend used symbols.}
    \label{fig:legends-and-symbols}
    \end{subfigure}
    
    \caption{The main models used in this work.}
    \label{fig:visual-model}
\end{figure*}

The visual model used in this work is similar to the one proposed in \cite{shen2019joint}. An Inception V3 model is used as the base model, followed by a global average 2D pooling layer, a dropout layer, and finally a fully connected layer with a  linear or softmax activation, depending on the task. The complete model is shown in Figure \ref{fig:visual-model}. The Inception V3 has been shown to perform well for the task of document quality assessment as shown in \cite{shen2019joint} and \cite{reddy2020nwqm}. Contrary to the approach used in these works, the model is trained using gradual unfreezing as explained in \ref{Gradual Unfreezing}, hence the name INCEPTION$_\text{GU}$ (INCEPTION gradual unfreezing).

Since it is too computationally expensive to load a large image dataset directly into memory, a generator is used which yields batches of images. No data augmentation is used since it does not make sense for scientific documents to be flipped, shifted, or rotated.

\subsection{Gradual Unfreezing}
\label{Gradual Unfreezing}
To effectively fine-tune a transfer learning model, or pre-trained model, it is important to apply techniques that prevent both catastrophic forgetting as well as slow convergence. In catastrophic forgetting, the pre-trained model loses all of its pre-training information due to large gradients in the initial training epochs.  With slow convergence, which is caused by fine-tuning the model too cautiously, the model can overfit without reaching an optimal solution. \cite{howard2018universal} propose a method called "gradual unfreezing", where layers are unfrozen iteratively during the training process, starting with the last layers which contain the most information from the pre-training task as shown in \cite{yosinski2014transferable}. It makes sense to first fine-tune these layers since the information held in these layers is the least transferable to the new task, in this case predicting accept/reject and citation labels for scientific documents. The other layers of the network are gradually unfrozen during the training process, until finally the entire pre-trained model is trained. Furthermore, the learning rate is reduced every time a new layer or set of layers is unfrozen to prevent the model from overfitting. 

The used method of gradual unfreezing is different for the accept/reject prediction and citation prediction tasks. For the citation prediction task, which is evaluated on the large ACL dataset, a method similar to the method proposed in \cite{howard2018universal} is used. Every few epochs, a new Inception block is unfrozen and the learning rate is lowered. Experiments show that applying this method for the accept/reject prediction task causes the model to overfit before an optimal solution is found. For this reason, the model is kept frozen for a longer time, replicating a feature extraction method. After the model starts to converge, all the Inception blocks are unfrozen simultaneously, after which the model is able to continue learning for a small number of epochs before overfitting. Details of the relevant hyperparameters are shown in Section \ref{Section:Hyper-parameters_and_Training_Techniques}.

\section{Joint Methods}
\label{Joint Methods}
In this section, the proposed joint model and its baselines are described. 

\subsection{Joint Baseline Models}
\label{Joint Baseline Models}
The Joint model is compared against the best-performing single models since any combination of textual and visual models could be combined into a joint model, which would be infeasible to test. The assumption is that the combination of the best-performing single models will also lead to the best joint model. The baseline models are thus selected based on the results shown in section \ref{subsection:main_results}. Furthermore, the results for the JOINT model from \cite{shen2019joint}, which is based on a combination of the \ac{BiLSTM} and INCEPTION models, are included for the accept/reject prediction task. 

\subsection{MultiSChuBERT}
\label{SChuBERTJOINT}

The joint model used in this work is based on a combination of the \ac{SChuBERT} model and the INCEPTION$_\text{GU}$ models. Since the main innovation is the \ac{SChuBERT} model, the model is named \ac{MultiSChuBERT}. The architecture of the \ac{MultiSChuBERT} model is shown in Figure \ref{fig:schubert-joint-model}. A document's visual grid and text are fed into the respective models. The textual model outputs a representation of size 256 while the visual model outputs a representation of size 2048. Since the used concatenation strategy expects both outputs to be of the same size, the visual output is passed through a fully connected layer with 256 hidden units with a ReLU activation. 

After this, the two outputs can be concatenated. As shown in \cite{reimers2019sentencebert}, the concatenation strategy used can have a large impact on the performance of the model. For the accept/reject prediction task, the concatenation strategy $(|u - v|)$ is used for the CL and LG subsets, and $(u, v, |u - v|)$ is used for the AI subset. For the citation prediction task, $(u, v, |u - v|,u * v)$ is used. These choices are based on the results of the relevant validation sets, for various experiments shown in section \ref{subsection:concat_results}. The concatenated result is passed through a fully connected layer with a ReLU activation and finally through a fully connected layer with a softmax or linear activation, depending on the task. The joint model uses a generator which is a combination of the textual and visual generators discussed before. A single example thus consists of the visual and textual inputs, which are both fed into the model.

\section{Multi-task Learning}
\label{Section:Multi-task-Learning}
Intuitively the accept-reject prediction task and citation-score prediction task are two related tasks that could potentially reinforce each other in a multi-task learning setup. Based on this intuition we have done experiments with multi-task learning, reported in Appendix \ref{Multi-task-Learning-Appendix}, however so far we have not obtained improvement using multi-task learning approaches.

\section{Hyperparameters and Training Techniques}
\label{Section:Hyper-parameters_and_Training_Techniques}
The used hyperparameters for the various models in this work are shown in table \ref{table:hyperparameters}. All hyperparameters are selected based on extensive experiments. A smaller learning rate is used for the accept/reject task than for the citation prediction task, as the PeerRead datasets are much smaller than the ACL dataset. 
The number of training epochs is the same for the accept/reject prediction and citation prediction tasks. However, for the INCEPTION$_\text{GU}$ and \ac{MultiSChuBERT} models, which both use gradual unfreezing, the number of sub-epochs is different. For accept/reject prediction, the model is trained for 30 epochs with the Inception model fully frozen, and then for 10 epochs with the model fully unfrozen. For citation prediction, the model is trained for 2 epochs for each of the 10 Inception blocks, and then 20 more epochs. 

During training, the model with the best validation score/loss for the designated selection metric is saved and used for prediction on the test set. Specifically, we select for number of citations prediction using the (highest) $R^{2}$ validation score, and for accept/reject prediction using the (lowest) binary cross-entropy validation loss. 
The learning rate is lowered every few epochs to prevent the model from overfitting.

\begin{table*}[t]
\centering
\scalebox{0.87}{
\begin{tabular}{llll}
\toprule
\cline{1-3}
\rowcolor[HTML]{FFFFFF} 
                                                                             & \textbf{SChuBERT}                                              & \textbf{INCEPTION$_\text{GU}$}                                           & \textbf{MultiSChuBERT}                                                                                           \\ \cline{1-3}
\midrule
\rowcolor[HTML]{D1D1D1} 
Vocabulary size                                                              & 30000                                                          & N/A                                                            & 30000                                                                                                                 \\
\rowcolor[HTML]{FFFFFF} 
Optimizer                                                                    & Adam                                                           & Adam                                                           & Adam                                                                                                                         \\
\rowcolor[HTML]{D1D1D1} 
Learning rate (AR, CIT)                                                      & 0.0001, 0.001                                                  & 0.0001, 0.001                                                  & 0.0001, 0.001                                                                                                               \\
\rowcolor[HTML]{FFFFFF} 
Epochs                                                                       & 40                                                             & 40                                                             & 40                                                                                                                             \\
\rowcolor[HTML]{D1D1D1} 
Loss function (AR, CIT)                                                      & CE, MAE                                                        & CE, MAE                                                        & CE, MAE                                                                                                                   \\
\rowcolor[HTML]{FFFFFF} 
Weight initialization                                                        & Xavier normal                                                  & Xavier normal                                                  & Xavier normal                                                                                                       \\
\rowcolor[HTML]{D1D1D1} 
Dropout rate (textual, visual)                                               & 0.3, N/A                                                       & N/A, 0.5                                                       & 0.3, 0.5                                                                                                                 \\
\rowcolor[HTML]{FFFFFF} 
GRU hidden size                                                              & 256                                                            & N/A                                                            & 256                                                                                                                           \\
\rowcolor[HTML]{D1D1D1} 
Joint hidden size                                                            & N/A                                                            & N/A                                                            & 128                                                                                                                           \\
\rowcolor[HTML]{FFFFFF}
\begin{tabular}[c]{@{}l@{}}Concatenation method (AR, CIT)\end{tabular}     & N/A                                                            & N/A                                                            & (u*v),  (u,v,|u-v|)                                                                                                           \\
\rowcolor[HTML]{D1D1D1} 
\begin{tabular}[c]{@{}l@{}}Train batch size (AI, CL, LG, ACL)\end{tabular} & \begin{tabular}[c]{@{}l@{}}18, 17, 17, 17\end{tabular} & \begin{tabular}[c]{@{}l@{}}18, 17, 17, 17\end{tabular} & \begin{tabular}[c]{@{}l@{}}18, 17, 17, 17\end{tabular} 
\\
\rowcolor[HTML]{FFFFFF}

\begin{tabular}[c]{@{}l@{}}Val batch size (AI, CL, LG, ACL)\end{tabular} & \begin{tabular}[c]{@{}l@{}}14, 16, 13, 15\end{tabular} & \begin{tabular}[c]{@{}l@{}}14, 16, 13, 15\end{tabular} & \begin{tabular}[c]{@{}l@{}}14, 16, 13, 15\end{tabular} 
\\
\rowcolor[HTML]{D1D1D1} 

\begin{tabular}[c]{@{}l@{}}Test batch size (AI, CL, LG, ACL)\end{tabular} & \begin{tabular}[c]{@{}l@{}}15, 13, 10, 18\end{tabular} & \begin{tabular}[c]{@{}l@{}}15, 13, 10, 18\end{tabular} & \begin{tabular}[c]{@{}l@{}}15, 13, 10, 18\end{tabular} 
\\
\rowcolor[HTML]{FFFFFF}

Word embedding size                                                          & 768                                                            & N/A                                                            & 768                                                                                                                           \\
\rowcolor[HTML]{D1D1D1} 
Image embedding size                                                         & N/A                                                            & 2048                                                           & 2048                                                               
\\
\bottomrule
\end{tabular}
}
\caption{Hyperparameters of the proposed models. 'AR, CIT' refers to the accept/reject prediction and citation prediction tasks. 'textual, visual' refers to the textual and visual portions of the joint model.}
\label{table:hyperparameters}
\end{table*}

A text chunk sequence length of 512 tokens without overlap is used
for all the \ac{SChuBERT}-based textual (sub)models. This choice is motivated by the results shown in Appendix \ref{seq_len_results}.

We repeat the experiments ten times, and report averaged scores and standard deviations over these repeated runs, to mitigate spurious random score differences resulting from different random model parameter initialization and other random factors.\footnote{The exception to this are the results for the CNN model, which were computed earlier on in the research, and results for the BiLSTM, HAN, HAN$_{ST}$ model, and SChuBERT* (earlier SChuBERT implementation) model which are literature results repeated from earlier work \citep{van_Dongen_2020,vanDongenMasterThesis}.
These models/experiments use an average over three, not ten runs.} 

\newchunk{Number of trainable parameters}

The different models differ in the number of total and trainable parameters. Table \ref{table:number_of_model_parameters} provides a summary, including a split between the number of trainable parameters in ``fully frozen'' and in ``fully unfrozen'' state, if applicable. 

First, note that the embedding model producing the text chunk encodings (e.g. BERT$_{\textrm{BASE}}$) is pre-trained, yet fully frozen during model training, and only the Gated Recurrent Unit and linear layer on top are trained. Consequently, in unfrozen state (if applicable) each of the visual and visual+textual models has at least 27 times more trainable parameters than SChuBERT, the purely textual model.
Furthermore, note that without gradual unfreezing, the number of trainable parameters matches the total number, since each parameter is set to be trainable. With gradual unfreezing, the final linear layer still is always trainable, and in the ``fully unfrozen state'' (only) all the Inception blocks are additionally unfrozen, not all parameters. This explains the different numbers of trainable parameters for the gradual unfreezing models in unfrozen state, and those for the corresponding models not using gradual unfreezing in the table. 

\begin{table*}
\caption{The numbers of total and trainable parameters for the different base models. }
\label{table:number_of_model_parameters}
\centering
\begin{tabular}{llll}
 \multirow{3}{*}{Model} & \multirow{1}{*}{\#total params}  & \multicolumn{2}{c}{\#trainable params} \\
 & & & \\
 &  & frozen & unfrozen \\
 \hline
 \EXCLUDE{ 
 \textbf{INCEPTION} & 24348900 & \multicolumn{2}{c}{24348900} \\
 \textbf{INCEPTION$_{GU}$} & 24348900 & 4098 & 21616994 \\
 \textbf{SChuBERT} & 788482  & \multicolumn{2}{c}{788482}\\
 \textbf{MultiSCHuBERT} & 25852640 & \multicolumn{2}{c}{25852640} \\
 \textbf{MultiSCHuBERT$_{GU}$} & 25852640 & 1312512 & 22925408 \\
 }
 \rowcolor[HTML]{d1d1d1}
 \textbf{SChuBERT} & 0.8M  & \multicolumn{2}{c}{0.8M}\\
 \textbf{INCEPTION} & 24.3M & \multicolumn{2}{c}{24.3M} \\
 \rowcolor[HTML]{d1d1d1}
 \textbf{INCEPTION$_{GU}$} & 24.3M & 4K & 21.6< \\
 \textbf{MultiSCHuBERT} & 25.9M & \multicolumn{2}{c}{25.9M} \\
 \rowcolor[HTML]{d1d1d1}
 \textbf{MultiSCHuBERT$_{GU}$} & 25.9M & 1.3M & 22.9M \\
 \hline
\end{tabular}
\end{table*}

\section{Experiments}
\label{Section:Experiments}
In this section, the various experiments used to answer the research question are described. 

\subsection{Multimodal Models for Accept/Reject Prediction and Citation Prediction}
\label{Accept/reject and citation prediction experiments}

In these main experiments, the focus is to research how to effectively apply multimodality in \ac{SDQP}, and in particular how to effectively train the MultiSCHuBERT model. To this end, the role of gradual unfreezing for effective training, part of \ref{RQ1},  is investigated.

For the task of accept/reject prediction, the PeerRead dataset is used for training and evaluation. For the citation prediction task, the ACL dataset is used. The various baseline models are trained and compared against the proposed textual and visual models. 

\newchunk{Earlier Experiments and Findings} 
Earlier work introduced the \ac{SChuBERT} model and applied it to the task of 
number of citations prediction \citep{van_Dongen_2020}. The model was shown to achieve substantial improvements over the \ac{BiLSTM} and \ac{HAN} baselines on this task. Other important conclusions from this work were: 1) The benefit of longer (i.e. full-text) input to the \ac{SChuBERT} model, 2) the benefit of more data to the model. 

Whereas the earlier work showed the performance advantage of full-text, even with a capped number of chunks such as to keep the input length of \ac{BERT} similar to that of the baseline models, \ac{SChuBERT} outperformed the baselines on the  citation prediction task. These experiments were further extended to accept/reject prediction, showing a similar trend of improved \ac{SChuBERT} results when using more chunks \cite{vanDongenMasterThesis}. These results are included in Appendix \ref{peerread_chunk_limit_experiments}.
As part of the master thesis research, it was also found that larger chunk length improves the results, and it is best to use no overlap between chunks. These experiments are included in Appendix
\ref{Sequence_Length_and_Overlap_Appendix} in this work.

We refer the reader to \citep{van_Dongen_2020, vanDongenMasterThesis} for more details on the earlier work.

\newchunk{Experiments in this work}
For the experiments reported here, we completely reimplemented the models in PyTorch and using PyTorch lightning. We also improved the preprocessing, in particular, we fixed some suboptimalities in the way the text was included from the parsed PDFs to form the paper text input. This led to improved results in comparison to the earlier implementation, for all models using the paper text, except the \ac{BiLSTM} and \ac{HAN} baselines which were using the same preprocessing already before.

\subsection{Concatenation Method}
\label{Concatenation Method}
 As shown in \cite{reimers2019sentencebert}, the used concatenation method can have a large impact on the performance of a joint model. 
The main focus of this research is combining visual and textual information; making the combining method an important aspect of the model. For this reason, multiple methods of concatenation are tested for both accept/reject prediction and citation prediction. These experiments thereby aim to investigate the role of the concatenation method in effective multimodal fusion, as part of  \ref{RQ1}.
 The following concatenation methods are tested, where $u$ and $v$ are the two embeddings to be combined:
 \begin{itemize}
     \item $(u, v)$: concatenation by taking $u$ and $v$ in one vector. 
     \item $(|u - v|)$: concatenation by taking the absolute element-wise difference between $u$ and $v$.
     \item $(u * v)$: concatenation by taking the element-wise product of $u$ and $v$. 
    \item $(u, v, |u - v|)$: concatenation of $u$, $v$ and their absolute element-wise difference. 
     \item $(u, v, u * v)$: concatenation of $u$, $v$ and their element-wise product.
\item $(u, v, |u - v|, u * v)$: concatenation of $u$, $v$, their absolute element-wise difference, and their element-wise product.
 \end{itemize}
 
 The goal of the concatenation layer is to capture relations between the embeddings. While it might be sufficient to simply add $v$ to $u$, it can be beneficial to add more types of information to the joint embedding by combining the textual and visual embeddings via a non-linear operation. \citet{conneau-etal-2017-supervised} and \citet{Cer2018UniversalSE} both use $(u, v, |u - v|, u * v)$, which seems to be the most complete concatenation method, but do not show results for other concatenation methods. \citet{reimers2019sentencebert} show that for their task, $(u, v, |u - v|)$ is the best-performing method, suggesting that more information is not necessarily better when it comes to combining embeddings.

\subsection{Pre-trained Text Embedding Specialized for the Science Domain}
\label{subsection:domain-specialized-bert-embedding}
In recent years, several embedding models have been proposed that are specifically trained for scientific documents. 
In these experiments, we try to answer \ref{RQ2}: 
\emph{Does multimodality still benefit the prediction when applying powerful pre-trained language models specialized for the science domain?}

These relevant scientific text embedding models use large collections of papers for training, either in full-text or using only title and abstract in addition to (optionally) using information from the citation graph. The additional embedding models we use are: 
\begin{enumerate}
 \item SciBERT \citep{Beltagy2019SciBERT}
 \item SciNCL \citep{SCINCL}
 \item SPECTER \cite{cohan2020specter} 
 \item SPECTER2.0 \cite{Singh2022SciRepEvalAM} 
\end{enumerate}

These models are different in the type of training data and training tasks they use. From an evaluation perspective, they also differ in the precision by which the data used for training these models is listed. Of these models, only for SPECTER2.0 is the description of the training data exact, and can the full set of examples used in training and validation of the model be readily downloaded as a set of json files. This is important for avoiding overlap between the data used for training the embedding model, and data used during testing, which in case of any overlap between the two amounts to \emph{label leakage}.

For SPECTER2, and only for that model, we can create data training and evaluation data that is guaranteed not to overlap with the data used for the training and validation of the embedding model itself. Using this information, we run three types of number of citation prediction experiments:

\newlist{enumexperiments}{enumerate}{2}
\setlist[enumexperiments,1]{label=Experiment \arabic*.,align=left}
\setlist[enumexperiments,2]{label=Experiment \arabic{enumstepsi}.\arabic*, align=left}

\begin{enumexperiments}
 \item Train number of citation prediction models using our original, full ACL training data, without change, but using the different domain-specialized embedding models.
 \item Use the trained models from the previous experiment. Then, for the baseline (BERT$_{\textrm{BASE}}$) and SPECTER2.0 models, evaluate on a new test set, filtered to exclude overlap with papers used in the training and validation of the SPECTER2.0 embedding model.
 \item Filter all the ACL data (training, validation, testing) to have no overlap with the SPECTER2.0 data, and use the filtered data to train and evaluate the models using the BERT$_{\textrm{BASE}}$ (baseline) and SPECTER2.0 embeddings.
\end{enumexperiments}

This way for all specialized embedding models we can test the performance as compared to BERT$_{\textrm{BASE}}$ and the other models (Experiment one), while not avoiding label leakage; and for SPECTER 2 we can furthermore test the effects while avoiding label leakage (Experiments two and three).

%% file: MainResultsTablePeerRead.tex
\begin{table*}[t]
\caption{Results for the visual (CNN, INCEPTION), textual (SChuBERT) and multimodal (MultiSChuBERT) models for the accept/reject prediction task.
}
\label{table:main_results_ar}
\centering
\begin{subfigure}{1.0\textwidth}
\centering
\caption{Results on the AI dataset.}

\scalebox{0.9}{
\mainresultstablePeerRead{
\midrule
\rowcolor[HTML]{d1d1d1}
\textbf{Maj Training Label} & 92.2 $\pm$ 0.00\%   &  0.500 $\pm$  0.000 & 0.000 $\pm$  0.00 \valscore{& 91.7 $\pm$ 0.00\%} & -- \\
\textbf{CNN} & 92.2 $\pm$ 0.00\%   & --  & -- \valscore{& --} & -- \\
\rowcolor[HTML]{d1d1d1}
\textbf{INCEPTION} &  92.3 $\pm$  1.36\% &  0.834 $\pm$  0.045 &  0.392 $\pm$  0.069 \valscore{&  92.6 $\pm$  1.20\%} & \epochNumberStartingFromOne{0.800} $\pm$  0.789\\ 
\textbf{INCEPTION$_{\textrm{GU}}$} &  93.0 $\pm$  0.87\% &  0.826 $\pm$  0.031 &  0.441 $\pm$  0.092 \valscore{&  92.5 $\pm$  0.95\%} &  \epochNumberStartingFromOne{30.500} $\pm$  0.707\\ 
\hline \hline
\rowcolor[HTML]{d1d1d1}
\textbf{SCHUBERT} &  93.5 $\pm$  0.52\% &  0.912 $\pm$  0.012 &  0.461 $\pm$  0.080 \valscore{&  91.9 $\pm$  0.35\%} &  \epochNumberStartingFromOne{18.200} $\pm$  5.534\\ 
\hline \hline
\textbf{MultiSChuBERT} &  92.7 $\pm$  0.43\% &  0.830 $\pm$  0.027 &  0.363 $\pm$  0.160 \valscore{&  92.7 $\pm$  1.15\%} &  \epochNumberStartingFromOne{0.900} $\pm$  0.876\\ 
\rowcolor[HTML]{d1d1d1}
\textbf{MultiSChuBERT$_{\textrm{GU}}$} &  \textbf{93.6 $\pm$  1.02\%} &  \textbf{0.913 $\pm$  0.020} &  \textbf{0.551 $\pm$  0.087} \valscore{&  \textbf{93.1 $\pm$  0.94}\%}  &  \epochNumberStartingFromOne{25.000} $\pm$  6.342\\ 
\bottomrule
}}
\end{subfigure}
\begin{subfigure}{1.0\textwidth}
\centering
\caption{Results on the CL dataset.}
\scalebox{0.9}{

\mainresultstablePeerRead{
\midrule
\rowcolor[HTML]{d1d1d1}
\textbf{Maj Training Label} & 68.9 $\pm$ 0.00\%   & 0.500 $\pm$  0.000 & 0.000 $\pm$  0.00 \valscore{&  78.0 $\pm$ 0.00\%} & -- \\
\textbf{CNN} & 68.9 $\pm$ 0.00\%  & --  & -- \valscore{& --} & -- \\
\rowcolor[HTML]{d1d1d1}
\textbf{INCEPTION} &  80.8 $\pm$  1.93\% &  0.871 $\pm$  0.020 &  0.667 $\pm$  0.072 \valscore{&  82.3 $\pm$  2.08\%} &  \epochNumberStartingFromOne{0.200} $\pm$  0.422\\ 
\textbf{INCEPTION$_{GU}$} &  80.2 $\pm$  3.38\% &  0.869 $\pm$  0.020 &  0.661 $\pm$  0.096 \valscore{&  \textbf{83.9 $\pm$  2.16\%}} &  \epochNumberStartingFromOne{31.000} $\pm$  1.633\\ 
\hline \hline
\rowcolor[HTML]{d1d1d1}
\textbf{SCHUBERT} &  82.4 $\pm$  2.14\% &  \textbf{0.920 $\pm$  0.004} &  0.640 $\pm$  0.070 \valscore{&  78.6 $\pm$  1.14\%} &  \epochNumberStartingFromOne{8.800} $\pm$  2.860\\ 
\hline \hline
\textbf{MultiSChuBERT} & 83.3 $\pm$  3.04\% & 0.893 $\pm$  0.023 &  0.708 $\pm$  0.099 \valscore{&  \textbf{83.9 $\pm$  1.56\%}} &  \epochNumberStartingFromOne{1.300} $\pm$  0.823\\ 
\rowcolor[HTML]{d1d1d1}
\textbf{MultiSChuBERT$_{\textrm{GU}}$} &  \textbf{85.2 $\pm$  1.20\%} &  \textbf{0.920 $\pm$  0.015} &  \textbf{0.740 $\pm$  0.032} \valscore{&  82.8 $\pm$  2.76\%} &  \epochNumberStartingFromOne{23.100} $\pm$  11.220\\
\bottomrule
}
}
\end{subfigure}
\begin{subfigure}{1.0\textwidth}
\centering
\caption{Results on the LG dataset.}
\scalebox{0.9}{

\mainresultstablePeerRead{
\midrule
\rowcolor[HTML]{d1d1d1}
\textbf{Maj Training Label} & 68.0 $\pm$  0.00\%  & 0.500 $\pm$  0.000 & 0.000 $\pm$  0.00 \valscore{& 63.5 $\pm$  0.00\%}  & -- \\
\textbf{CNN} & 65.7 $\pm$ 2.79\%   & -- & -- \valscore{& --} & -- \\
\rowcolor[HTML]{d1d1d1}
\textbf{INCEPTION} &  82.2 $\pm$  1.42\% &  0.904 $\pm$  0.011 &  0.729 $\pm$  0.026 \valscore{&  83.3 $\pm$  2.52\%} &  \epochNumberStartingFromOne{1.500} $\pm$  2.121\\ 
\textbf{INCEPTION$_\textrm{GU}$} &  83.6 $\pm$  1.86\% &  0.904 $\pm$  0.013 &  0.752 $\pm$  0.023 \valscore{&  84.1 $\pm$  1.45\%} &  \epochNumberStartingFromOne{30.600} $\pm$  0.516\\ 
\hline \hline
\rowcolor[HTML]{d1d1d1}
\textbf{SCHUBERT} &  80.3 $\pm$  1.37\% &  0.880 $\pm$  0.006 &  0.723 $\pm$  0.014 \valscore{&  76.9 $\pm$  0.56\%} &  \epochNumberStartingFromOne{12.000} $\pm$  3.091\\ 
\hline \hline
\textbf{MultiSChuBERT} &  83.4 $\pm$  1.65\% &  0.921 $\pm$  0.012 &  0.750 $\pm$  0.017 \valscore{&  \textbf{84.9 $\pm$  1.80\%}} &  \epochNumberStartingFromOne{0.900} $\pm$  0.876\\ 
\rowcolor[HTML]{d1d1d1}
\textbf{MultiSChuBERT$_{\textrm{GU}}$}  &  \textbf{84.9 $\pm$  1.40\%} &  \textbf{0.931 $\pm$  0.007} &  \textbf{0.781 $\pm$  0.016} \valscore{&  83.5 $\pm$  1.56\%} &  \epochNumberStartingFromOne{31.300} $\pm$  1.947\\ 
\bottomrule
}}
\end{subfigure}
\end{table*}

\begin{table*}[t]
\caption{Comparison between accuracy results for our MultiSChuBERT$_\text{GU}$ model and literature results on the accept/reject prediction task.
SChuBERT* denotes results reported for the original SChuBERT implementation \citep{vanDongenMasterThesis}; whereas BiLSTM and HAN$_{\textrm{ST}}$ \citep{structure_tags} are two other textual models and the JOINT model \citet{shen2019joint} is another multimodal model.}
\label{table:comparison_literature_results_ar}
\centering
\scalebox{0.90}{
\begin{tabular}{lllll|ll}
\toprule
\multirow{2}{*}{\textbf{Dataset}} & \multicolumn{4}{c}{Literature models \& results} & \multicolumn{2}{c}{Our models \& results}  \\
& \multirow{2}{*}{\textbf{BiLSTM}}   & \multirow{2}{*}{\textbf{HAN$_\text{ST}$}} & \multirow{2}{*}{\textbf{JOINT}} & \multirow{2}{*}{\textbf{SChuBERT*}} & \multirow{2}{*}{\textbf{SChuBERT}}   & \multirow{2}{2cm}{\textbf{Multi- SChuBERT$_\text{GU}$}}          \\ 
 & & & & & &  \\
\midrule
\rowcolor[HTML]{d1d1d1} 

AI      & 91.5 $\pm$ 1.03\%   & 89.6 $\pm$ 1.02\% & \textbf{93.4 $\pm$ 1.07\%}  & 92.4 $\pm$ 0.84\% & 93.5 $\pm$  0.52\%   & \textbf{93.6 $\pm$  1.02\%} \\

CL      & 76.2 $\pm$ 1.30\%  & 81.8 $\pm$ 1.91\%  & 77.1 $\pm$ 3.10\% & 80.8 $\pm$ 2.60\% & 82.4 $\pm$  2.14\%  & \textbf{85.2 $\pm$  1.20\%} \\ 

\rowcolor[HTML]{d1d1d1} 

LG      & 81.1 $\pm$ 0.83\%  & 78.7 $\pm$ 0.69\% & 79.9 $\pm$ 2.54\% &  80.2  $\pm$ 1.39\% & 80.3 $\pm$  1.37\%  & \textbf{84.9 $\pm$  1.40\%} \\ 

\bottomrule
\end{tabular}
}
\label{table:joint_ar}
\end{table*}

\begin{table*}[t]
\begin{minipage}{1.0\textwidth}
\caption{Results for the visual (CNN, INCEPTION), textual (SChuBERT) and multimodal (MultiSChuBERT) models for the citation prediction task.
SChuBERT* denotes results reported for the original SChuBERT implementation, whereas BiLSTM and HAN$_{\textrm{ST}}$ \citep{structure_tags} are two other textual models; these results are repeated from \citep{van_Dongen_2020}.}
\label{table:main_results_citation}
\centering
\scalebox{0.9}{
\mainresultstableACL{
\rowcolor[HTML]{d1d1d1} 
 \textbf{Avg Training Label} & -0.005 $\pm$  0.000 &  1.028 $\pm$  0.000 &  1.643 $\pm$  0.000 \valscore{& -0.001 $\pm$  0.000} &  -- \\
 \textbf{BiLSTM} &  0.319  $\pm$ 0.013  & 1.110 $\pm$  0.021 &  0.824 $\pm$ 0.009 & -- &  -- \\
\rowcolor[HTML]{d1d1d1} 
 \textbf{HAN} & 0.339 $\pm$ 0.013 &  1.080 $\pm$  0.021 &   0.820 $\pm$ 0.009  & -- &  -- \\ 
  \textbf{SChuBERT*} &  0.398 $\pm$  0.006 & 0.985 $\pm$   0.010 & 0.789 $\pm$ 0.005 &  -- &  -- \\
\rowcolor[HTML]{d1d1d1}
\textbf{CNN} &0.118 $\pm$ 0.009 & 1.444 $\pm$ 0.013 & 0.952 $\pm$ 0.003 \valscore{& --} & -- \\
  \hline \hline
\textbf{INCEPTION} &  0.275 $\pm$  0.029 &  1.186 $\pm$  0.048 &  0.852 $\pm$  0.018  \valscore{&  0.265 $\pm$  0.016} &  \epochNumberStartingFromOne{7.700} $\pm$  3.302\\ 
 \rowcolor[HTML]{d1d1d1}
 \textbf{INCEPTION$_{\textrm{GU}}$} &  0.332 $\pm$  0.014 &  1.092 $\pm$  0.023 &  0.786 $\pm$  0.009 \valscore{&  0.329 $\pm$  0.011} &  \epochNumberStartingFromOne{37.400} $\pm$  2.413\\ 
 \hline \hline
\textbf{SCHUBERT} &  0.432 $\pm$  0.010 &  0.929 $\pm$  0.017 &  0.765 $\pm$  0.009  \valscore{&  0.394 $\pm$  0.005} &  \epochNumberStartingFromOne{22.300} $\pm$  8.512\\ 
\hline \hline
\rowcolor[HTML]{d1d1d1}
\textbf{MultiSChuBERT} &  0.427 $\pm$  0.016 &   0.937 $\pm$  0.026 &  0.760 $\pm$  0.009 \valscore{&  0.409 $\pm$  0.010} &  \epochNumberStartingFromOne{12.700} $\pm$  6.499\\ 
\textbf{MultiSChuBERT$_\text{GU}$} &  \textbf{0.454 $\pm$  0.006} &  \textbf{0.893 $\pm$  0.010} &  \textbf{0.717 $\pm$  0.006} \valscore{&  \textbf{0.436 $\pm$  0.012}} &  \epochNumberStartingFromOne{36.600} $\pm$  2.221\\ 
\bottomrule
}}
 
\end{minipage}

\end{table*}

%% file: EvaluationMetrics.tex
\section{Evaluation Metrics}
\label{Section:Evaluation_Metrics}
Choosing the right evaluation metric for the experiments is important. For the accept/reject prediction task,  \emph{accuracy} is used as the metric, as it is easy to interpret, and because baseline models from other papers also use accuracy as the main evaluation metric. Therefore, using this metric makes our results comparable to other work in the literature.  
In addition to accuracy, we also report the \emph{ROC AUC} and \emph{$F_{1}$-score} metrics. These two metrics for binary classification are often used in scenarios where there is a considerable imbalance in the frequency of the two classes, which is clearly the case for the PeerRead dataset, see Table \ref{datasets-classbalance}. Note that whereas ROC AUC is ``insensitive'' to class imbalance, this allows it to mask poor performance for recognition of the positive class in cases where there is a large class imbalance, with a higher ratio of negative instances. Since the false positive rate is pulled down by the large number of true negatives in such scenarios \cite{inbalancedDataMetricsUseRecommendations, precisionRecallCurveMoreInformativeThanROC}.

Regression problems such as citation prediction involve the prediction of continuous values. As predictions almost never exactly match the target, the choice for evaluation is how to score differences from the target.
The most common evaluation metrics are \ac{MSE} and \ac{MAE} \cite{knight1999mathematical}.
While good at comparing the performance of different models, these metrics do not directly explain whether the model is good in general. For this reason, in addition to the \ac{MSE} and \ac{MAE} metrics, the
R$^2$ (coefficient of determination) metric is used \citep{devore2011modern}.
%
The R$^2$ metric divides the \ac{MSE} of the predictions by the variance for the labels. It measures how much of the dependent variable can be explained by the model. A R$^2$ score of 0 means that the model always predicts the average target label, while an R$^2$ score of 1 means that the model can perfectly predict the data. This property makes it more interpretable than the MSE and MAE, for which the values can vary highly based on the dataset.

%% file: ResultsPyTorchReimplementation.tex
\section{Results}
\label{Section:Results}
This section describes the results for the different models and experiments. The main results in which the MultiSChuBERT model is compared against the other baseline model, using BERT$_{\textrm{BASE}}$ embeddings, 
are shown in subsection \ref{subsection:main_results}. The results for comparing the different concatenation methods for the MultiSChuBERT model are shown in  subsection \ref{subsection:concat_results}.
Finally, the results for the use of domain-specialized embeddings in combination with SChuBERT and MultiSChuBERT are shown in subsection \ref{subsection:domain_specialized_embeddings_results}.

\subsection{Main results}
\label{subsection:main_results}

\newchunk{Accept/reject prediction}
\label{main_accept_reject_results}
The main results for the accept/reject prediction task are shown in Table \ref{table:main_results_ar}. 
The MultiSChuBERT$_{GU}$ model performs best, outperforming all other models.

The INCEPTION$_\text{GU}$ model is able to outperform the Inception model from \cite{shen2019joint} on two out of three datasets. MultiSChuBERT$_{GU}$  outperforms MultiSChuBERT for all three datasets.  This shows the benefit and importance of gradually unfreezing layers the in pre-trained models.

The CNN model is not able to improve upon the majority baseline. For the LG dataset, the CNN even performs under the majority baseline. This shows the importance of a strong enough and pre-trained model visual model, especially in the setting of small data sizes of PeerRead data.  

Comparing to results in the literature, Table \ref{table:comparison_literature_results_ar} shows a comparison of the MultiSChuBERT$_{GU}$ model against other baseline models from the literature, showing improvement over all these models, including markedly JOINT 
\cite{shen2019joint}, the earlier multimodal model for this task.

\newchunk{Citation prediction}
\label{main_citation_results}
Results for comparing the MultiSChuBERT models to the other models are shown in Table \ref{table:main_results_citation}.
First, it can be seen that on this task, the textual and multimodal models work much better than the visual models. MultiSCHuBERT$_{GU}$, the best performing model is able to further improve the performance on the ACL dataset. Second, gradual unfreezing helps in all cases for this dataset. Finally, while still not perfect, there is increased coherence between validation and test scores on this dataset.

\newchunk{Analysis}
To fully understand the observed effect of gradual unfreezing, it is important to recall its implementation in the PeerRead setting. Here, for the visual models, 
it amounts to only training the linear output layer for the first thirty epochs and then unfreezing all Inception blocks at once. For the multimodal models, gradual unfreezing implies the same for the INCEPTION sub-model, but here there is also the textual sub-model which is fully trained throughout.
This can be thought of as giving the textual model, which has much less parameters and is therefore likely slower to fit the training data, a ``head-start'' during training.   

In both cases the INCEPTION (sub-)model quickly gets overfitted, and gradual unfreezing helps against this. However, in the multimodal case gradual unfreezing offers added benefit of providing time for the textual model to be trained without the problem of overfitting the  INCEPTION (sub-)model in the meantime; which explains the larger impact in this setting.  

A mismatch between the validation and test results is another notable thing. In particular, the validation scores for SChuBERT and MultiSChuBERT$_{GU}$ give an underestimate of the test scores, while those for INCEPTION, INCEPTION$_{GU}$ and MultiSChuBERT give an overestimate. We believe there are two factors that could explain this mismatch, and how it plays out for different models:

\begin{enumerate}
 \item \textit{The ratio between accept and reject labels, which differs substantially between the PeerRead training, validation and test sets.} Consequently, following the training data statistics in terms of this ratio will give quite different results for the validation vs test set. This effect can be seen in the result rows for the Majority Training Label Prediction (Maj Training Label).
 \item \textit{Differences in the number of trainable parameters between the (unfrozen) INCEPTION (sub-)model and the SChuBERT (sub-)model.} Noting that since the INCEPTION model has much more trainable parameters than SCHuBERT (see Table \ref{table:number_of_model_parameters}), it can be expected to be much more prone to overfit the training data, and this is indeed observed during training.  In this context, it has also been noted in the literature \cite{JMLR:v11:cawley10a} that models with substantially more trainable parameters can not only overfit the training data, but also the validation data more, and that particular for small training and validation sets this can cause problems.

\end{enumerate}

The last column of Table \ref{table:main_results_ar} shows the epoch numbers of the validation-set-selected models. These show that compared to the other models, SChuBERT and MultiSChuBERT$_{GU}$ are the only ones that either use only the textual model or train it for a substantial number of epochs (24.1 or more). In contrast, the MultiSChuBERT model cannot be expected to give as much weight to the textual sub-model, since it ends up training it for only 2.3 epochs or less.
We thus note that INCEPTION, INCEPTION$_{GU}$ and MultiSChuBERT each rely only or at least more strongly on the more overfitting-prone visual (sub-)model. This would explain their observed tendency to perform high on the validation data, while not realizing the same performance on the test set.\\

Fortunately, the discussed challenges with the results on the PeerRead data are mostly peculiarities due to the small and unbalanced nature of this dataset and its subsets. As we observed, these mostly disappear on the larger ACL dataset for number of citations prediction.

\begin{table*}[t]
\caption{MultiSChuBERT$_\text{GU}$ results for the different concatenation methods, for the citation prediction task, evaluated on the ACL dataset.
}

\centering
\scalebox{0.9}{
\begin{tabular}{cccc||cc}
\hline
\multirow{2}{1.2cm}{concatenation method} & \multicolumn{3}{c}{test scores} & \multicolumn{2}{c}{validation scores \& statistics}\\
& R2$\uparrow$ & MSE$\downarrow$ & MAE$\downarrow$ & R2$\uparrow$ & model epoch \\
\midrule
\rowcolor[HTML]{d1d1d1}
$(u, v)$ &  0.446 $\pm$  0.010 &  0.905 $\pm$  0.016 &  0.723 $\pm$  0.006 &     0.431 $\pm$  0.005 & \epochNumberStartingFromOne{36.700} $\pm$  1.160\\  
$(|u - v|)$ &  0.449 $\pm$  0.007 &  0.901 $\pm$  0.012 &  0.722 $\pm$  0.006  &  0.429 $\pm$  0.008 & \epochNumberStartingFromOne{37.000} $\pm$  3.266\\ 
\rowcolor[HTML]{d1d1d1}
$(u * v)$ &  0.443 $\pm$  0.013 &  0.910 $\pm$  0.021 &  0.731 $\pm$  0.016  &  0.431 $\pm$  0.006 & \epochNumberStartingFromOne{34.400} $\pm$  8.708\\ 
$(|u - v|, u * v)$ &  0.442 $\pm$  0.011 &  0.912 $\pm$  0.019 &  0.726 $\pm$  0.008 &  0.424 $\pm$  0.006  & \epochNumberStartingFromOne{37.200} $\pm$  2.440\\ 
\rowcolor[HTML]{d1d1d1}
$(u,v,u * v)$ &  0.445 $\pm$  0.010 &  0.908 $\pm$  0.017 &  0.725 $\pm$  0.008 &  0.433 $\pm$  0.007 &  \epochNumberStartingFromOne{36.900} $\pm$  2.424 \\ 
$(u, v, |u - v|)$ &  0.450 $\pm$  0.005 &  0.900 $\pm$  0.009 &  0.721 $\pm$  0.006 &  \textbf{0.436 $\pm$  0.009}  & \epochNumberStartingFromOne{37.100} $\pm$  2.998  \\ 
\rowcolor[HTML]{d1d1d1}
$(u, v, |u - v|,u * v)$ &  \textbf{0.454 $\pm$  0.006} &  \textbf{0.893 $\pm$  0.010} &  \textbf{0.717 $\pm$  0.006}  &  \textbf{0.436 $\pm$  0.012} &  \epochNumberStartingFromOne{36.600} $\pm$  2.221\\ 
\bottomrule
\end{tabular}
}
\label{table:concatenation_cit}

\end{table*}

\begin{table*}[t]
\caption{MultiSChuBERT$_\text{GU}$ results for the different concatenation methods, for the accept/reject prediction task, evaluated on the PeerRead datasets.}

\centering

\begin{subfigure}{1.0\textwidth}
\caption{Results on the AI dataset.}
\centering
\scalebox{0.9}{
\begin{tabular}{cccc||cc}
\hline
\multirow{2}{1cm}{concatenation method} & \multicolumn{3}{c}{test scores} & \multicolumn{2}{c}{validation scores \& statistics}\\
 & Accuracy & ROC$\uparrow$ AUC$\uparrow$ & $F_{1}$-score$\uparrow$ & Accuracy$\uparrow$  & model epoch \\
\midrule
\rowcolor[HTML]{d1d1d1}
$(u, v)$ &  93.9 $\pm$  0.70\% &  \textbf{0.922 $\pm$  0.012} &  \textbf{0.578 $\pm$  0.055} &  92.9 $\pm$  0.70\% &  \epochNumberStartingFromOne{25.800} $\pm$  2.741\\ 
$(|u - v|)$ &  93.7 $\pm$  0.61\% &  0.912 $\pm$  0.009 &  0.506 $\pm$  0.076 &  92.4 $\pm$  0.78\% &  \epochNumberStartingFromOne{16.500} $\pm$  6.078\\ 
\rowcolor[HTML]{d1d1d1}
$(u * v)$ &  93.5 $\pm$  0.69\% &  0.894 $\pm$  0.008 &  0.481 $\pm$  0.055 &  92.2 $\pm$  0.75\% &  \epochNumberStartingFromOne{15.300} $\pm$  4.473\\ 
$(|u - v|, u * v)$ &  \textbf{94.0 $\pm$  0.52\%} &  0.907 $\pm$  0.015 &  0.533 $\pm$  0.086 &  92.4 $\pm$  0.66\% &  \epochNumberStartingFromOne{17.000} $\pm$  7.916\\ 
\rowcolor[HTML]{d1d1d1}
$(u,v,u * v)$ &  93.7 $\pm$  0.78\% &  0.908 $\pm$  0.012 &  0.484 $\pm$  0.122 &  92.1 $\pm$  0.71\% &  \epochNumberStartingFromOne{15.500} $\pm$  5.759\\ 
$(u, v, |u - v|)$ &  93.6 $\pm$  1.02\% &  0.913 $\pm$  0.020 &  0.551 $\pm$  0.087 &  \textbf{93.1 $\pm$  0.94}\% &  \epochNumberStartingFromOne{25.000} $\pm$  6.342\\ 
\rowcolor[HTML]{d1d1d1}
$(u, v, |u - v|,u * v)$ &  93.8 $\pm$  1.08\% &  0.909 $\pm$  0.016 &  0.488 $\pm$  0.165 &  92.5 $\pm$  0.87\% &  \epochNumberStartingFromOne{15.200} $\pm$  7.671\\ 
\bottomrule
\end{tabular}
}
\end{subfigure} \\

\vspace{0.3cm}
\begin{subfigure}{1.0\textwidth}
\caption{Results on the CL dataset.}
\centering
\scalebox{0.9}{
\begin{tabular}{cccc||cc}
\hline
\multirow{2}{1cm}{concatenation method} & \multicolumn{3}{c}{test scores} & \multicolumn{2}{c}{validation scores \& statistics}\\
 & Accuracy$\uparrow$ & ROC AUC$\uparrow$ & $F_{1}$-score$\uparrow$ & Accuracy$\uparrow$  & model epoch \\
\midrule
\rowcolor[HTML]{d1d1d1}
$(u, v)$ &  84.9 $\pm$  2.03\% &  0.917 $\pm$  0.005 &  0.733 $\pm$  0.053 &  81.8 $\pm$  2.74\% &  \epochNumberStartingFromOne{21.400} $\pm$  11.918\\ 
$(|u - v|)$ &  85.2 $\pm$  1.20\% &  0.920 $\pm$  0.015 &  0.740 $\pm$  0.032 &  \textbf{82.8 $\pm$  2.76}\% &  \epochNumberStartingFromOne{23.100} $\pm$  11.220\\ 
\rowcolor[HTML]{d1d1d1}
$(u * v)$ &  85.5 $\pm$  1.37\% &  \textbf{0.921 $\pm$  0.007} &  0.742 $\pm$  0.037 &  78.9 $\pm$  0.54\% &  \epochNumberStartingFromOne{7.000} $\pm$  1.247\\ 
$(|u - v|, u * v)$ &  \textbf{85.8 $\pm$  1.24}\% &  0.918 $\pm$  0.014 &  \textbf{0.758 $\pm$  0.023} &  80.2 $\pm$  1.85\% &  \epochNumberStartingFromOne{16.900} $\pm$  11.949\\ 
\rowcolor[HTML]{d1d1d1}
$(u,v,u * v)$ &  85.4 $\pm$  1.96\% &  0.918 $\pm$  0.008 &  0.749 $\pm$  0.048 &  79.8 $\pm$  2.97\% &  \epochNumberStartingFromOne{11.700} $\pm$  10.166\\ 
$(u, v, |u - v|)$ &  \textbf{85.8 $\pm$  2.40}\% &  0.919 $\pm$  0.010 &  0.755 $\pm$  0.052 &  81.8 $\pm$  2.97\% &  \epochNumberStartingFromOne{22.200} $\pm$  11.708\\ 
\rowcolor[HTML]{d1d1d1}
$(u, v, |u - v|,u * v)$ &  \textbf{85.8 $\pm$  1.88}\% & 
\textbf{0.921 $\pm$  0.006} &  0.747 $\pm$  0.050 &  80.5 $\pm$  2.81\% &  \epochNumberStartingFromOne{15.000} $\pm$  11.235\\ 
\bottomrule
\end{tabular}}
\end{subfigure} \\

\vspace{0.3cm}
\begin{subfigure}{1.0\textwidth}
\caption{Results on the LG dataset.}
\centering
\scalebox{0.9}{
\begin{tabular}{cccc||cc}
\hline
\multirow{2}{1cm}{concatenation method} & \multicolumn{3}{c}{test scores} & \multicolumn{2}{c}{validation scores \& statistics}\\
 & Accuracy$\uparrow$ & ROC AUC$\uparrow$ & $F_{1}$-score$\uparrow$ & Accuracy$\uparrow$  & model epoch \\
\midrule
\rowcolor[HTML]{d1d1d1}
$(u, v)$ &  84.2 $\pm$  2.02\% &  0.924 $\pm$  0.009 &  0.762 $\pm$  0.028 &  83.2 $\pm$  1.91\% &  \epochNumberStartingFromOne{31.500} $\pm$  0.972\\ 
$(|u - v|)$ &  \textbf{84.9 $\pm$  1.40}\% &  \textbf{0.931 $\pm$  0.007} &  \textbf{0.781 $\pm$  0.016} &  \textbf{83.5 $\pm$  1.56}\% &  \epochNumberStartingFromOne{31.300} $\pm$  1.947\\ 
\rowcolor[HTML]{d1d1d1}
$(u * v)$ &  81.8 $\pm$  1.87\% &  0.908 $\pm$  0.007 &  0.725 $\pm$  0.033 &  81.4 $\pm$  2.36\% &  \epochNumberStartingFromOne{25.800} $\pm$  6.070\\ 
$(|u - v|, u * v)$ &  82.6 $\pm$  1.68\% &  0.912 $\pm$  0.010 &  0.750 $\pm$  0.020 &  81.2 $\pm$  3.11\% &  \epochNumberStartingFromOne{25.900} $\pm$  6.008\\ 
\rowcolor[HTML]{d1d1d1}
$(u,v,u * v)$ &  83.6 $\pm$  1.88\% &  0.918 $\pm$  0.009 &  0.760 $\pm$  0.020 &  82.0 $\pm$  3.36\% &  \epochNumberStartingFromOne{26.700} $\pm$  7.931\\ 
$(u, v, |u - v|)$ &  84.2 $\pm$  1.59\% &  0.921 $\pm$  0.013 &  0.767 $\pm$  0.027 &  82.7 $\pm$  2.73\% &  \epochNumberStartingFromOne{29.400} $\pm$  4.624\\ 
\rowcolor[HTML]{d1d1d1}
$(u, v, |u - v|,u * v)$ &  82.5 $\pm$  1.15\% &  0.912 $\pm$  0.009 &  0.750 $\pm$  0.016 &  81.7 $\pm$  1.96\% &  \epochNumberStartingFromOne{27.300} $\pm$  6.550\\ 
\bottomrule
\end{tabular}
}
\end{subfigure}
\label{table:concatenation_ar}
\hspace{0.3cm}
\end{table*}

\subsection{Concatenation method results}
\label{subsection:concat_results}
The results for the concatenation experiments are shown in Table \ref{table:concatenation_cit} (citation prediction) and Table \ref{table:concatenation_ar} (accept/reject prediction). 
For the citation prediction task, all concatenation methods outperform the best performing textual and visual models. The concatenation method that performs the best on the test set is $(u, v, |u - v|,u * v)$, and this method also achieves the highest validation accuracy (shared with one other concatenation method). 

For the accept/reject prediction task, there is no clear best performing concatenation method. Even so, choosing the concatenation method giving the best performance on the validation set gives a MultiSChuBERT system that that does outperform all other baseline models, see Table \ref{table:main_results_ar}. 

\newchunk{Analysis}

 It can be noted that the most complete concatenation method $(u, v, |u - v|,u * v)$ does give the best results on the number of citations prediction task, which uses the relatively large ACL dataset.
 In contrast, on the much smaller PeerRead data subsets for the different domains, there is no clear winner, and simpler concatenation methods often seem to work better. On the LG PeerRead subset, there is coherence between the best performing method on the validation and test sets ($|u-v|$). This dataset is both larger and markedly more balanced than the two other PeerRead subsets (AI and CL). This suggests that provided the availability of sufficient and for classification somewhat balanced data, the choice of the concatenation method makes a difference for the model performance and can be predicted from the validation set. Also, with substantially more data (ACL) where the risk of overfitting is much reduced, the more complex/complete concatenation method benefits. 
 Finally, looking again at the results in  Table \ref{table:main_results_ar}, the use of gradual unfreezing is an even more important factor for performance than the choice of concatenation method. This can be understood from the fact that this factor strongly contributes to the emergence of a balanced fusion between modalities during learning, particularly in scenarios where one the sub-models has much more trainable parameters then the other when fully unfrozen; as in the case of the INCEPTION and SChuBERT sub-models
 (see Table \ref{table:number_of_model_parameters}).

\begin{table*}
 \caption{Label statistics of the original and filtered ACL datasets.}
 \label{table:ACL_datasets_label_statistics}
 \begin{subtable}[b]{0.48\textwidth}
 \caption{ACL data}
 \scalebox{0.8}{
 \begin{tabular}{cccc}
  \midrule
  \rowcolor[HTML]{d1d1d1} 
  subset & train & val & test  \\
  \#examples &  27852 &  1547 & 1548  \\
  \rowcolor[HTML]{d1d1d1} 
  avg label & 1.729$\pm$1.191 & 1.759$\pm$1.216 & 1.819$\pm$1.279  \\ 
  \bottomrule
 \end{tabular}  
 }
 \end{subtable}
 \hspace{0.5cm}
 \begin{subtable}[b]{0.48\textwidth}
 \caption{Filtered ACL data}
 \scalebox{0.8}{
  \begin{tabular}{cccc}
  \midrule
  \rowcolor[HTML]{d1d1d1} 
  subset & train & val & test  \\
  \#examples &  16730 & 957 &   926\\
  \rowcolor[HTML]{d1d1d1} 
  avg label & 1.330$\pm$0.978 & 1.350$\pm$0.991 & 1.360$\pm$1.023 \\
  \bottomrule
 \end{tabular} 
 }
 \end{subtable}
\end{table*}

\begin{table*}
\centering
\caption{{
  Results citation task evaluated using the ACL data and different types of embeddings for the textual model (BERT$_{\textrm{BASE}}$, SciBERT, scincl, SPECTER, SPECTER2.0). If not specified otherwise, BERT$_{\textrm{BASE}}$ embeddings are used.
}
}
\label{table:different-embedding-results-original-data}

\scalebox{0.88}{
\mainresultstableACL{
 \rowcolor[HTML]{d1d1d1} 
  \textbf{Avg Training Label} & -0.005 $\pm$  0.000 &  1.028 $\pm$  0.000 &  1.643 $\pm$  0.000 \valscore{& -0.001 $\pm$  0.000} &  -- \\ 
 \textbf{SChuBERT} &  0.432 $\pm$  0.010 &  0.765 $\pm$  0.009 &  0.929 $\pm$  0.017 \valscore{&  0.394 $\pm$  0.005} &  \epochNumberStartingFromOne{22.300} $\pm$  8.512\\ 
 \rowcolor[HTML]{d1d1d1} 
 \textbf{MultiSChuBERT$_{\textrm{GU}}$} &  0.454 $\pm$  0.006 &  0.717 $\pm$  0.006 &  0.893 $\pm$  0.010 \valscore{&  0.436 $\pm$  0.012} &  \epochNumberStartingFromOne{36.600} $\pm$  2.221\\
 \hline \hline
 \textbf{SChuBERT$_{\textrm{SCIBERT}}$} &  0.467 $\pm$  0.014 &  0.743 $\pm$  0.011 &  0.871 $\pm$  0.022 \valscore{&  0.439 $\pm$  0.005} &  \epochNumberStartingFromOne{14.600} $\pm$  3.658\\ 
 \rowcolor[HTML]{d1d1d1} 
 \textbf{SChuBERT$_{\textrm{SCINCL}}$} &  0.460 $\pm$  0.008 &  0.751 $\pm$  0.006 &  0.883 $\pm$  0.013 \valscore{&  0.447 $\pm$  0.006} &  \epochNumberStartingFromOne{32.300} $\pm$  5.478\\ 
  \textbf{SChuBERT$_{\textrm{SPECTER}}$} &  0.447 $\pm$  0.013 &  0.754 $\pm$  0.010 &  0.904 $\pm$  0.021 \valscore{&  0.440 $\pm$  0.009} &  \epochNumberStartingFromOne{23.700} $\pm$  10.144\\ 
  \rowcolor[HTML]{d1d1d1} 
 \textbf{SChuBERT$_{\textrm{SPECTER2.0}}$} &  0.474 $\pm$  0.013 &  0.736 $\pm$  0.009 &  0.860 $\pm$  0.021 \valscore{&  0.460 $\pm$  0.003} &  \epochNumberStartingFromOne{13.400} $\pm$  6.186\\ 
& & & \valscore{&} &  \\
 \multirow{-2}{3.9cm}{\textbf{Multi- SChuBERT$_{\textrm{GU\_SPECTER2.0}}$}} &  \multirow{-2}{*}{\textbf{0.503 $\pm$  0.011}} &  \multirow{-2}{*}{\textbf{0.693 $\pm$  0.016}} &  \multirow{-2}{*}{\textbf{0.813 $\pm$  0.018}} \valscore{&  \multirow{-2}{*}{\textbf{0.484 $\pm$  0.009}}} &  \multirow{-2}{*}{\epochNumberStartingFromOne{31.300} $\pm$  11.898}\\ 
 \bottomrule
}
}
\end{table*}

\begin{table*}
\centering
\caption{SPECTER2.0 results citation task, original ACL training and validation data and filtered test set. 
Validation scores and statistics are equivalent to those of the matching systems from Table \ref{table:different-embedding-results-original-data}, but are repeated for ease of comparison.}
\label{table:different-embedding-results-original-data-filtered-testset}

\scalebox{0.88}{
\mainresultstableACL{
\midrule
\rowcolor[HTML]{d1d1d1} 
\textbf{Avg Training Label} & -0.130 $\pm$  0.000 &  0.910 $\pm$  0.000 &  1.181 $\pm$  0.000 & -0.001 $\pm$  0.000 &  --\\ 
\textbf{SChuBERT} &  0.267 $\pm$  0.015 &  0.693 $\pm$  0.009 &  0.766 $\pm$  0.015 \valscore{&  0.394 $\pm$  0.005} &  \epochNumberStartingFromOne{22.300} $\pm$  8.512\\ 
\rowcolor[HTML]{d1d1d1} 
 \textbf{MultiSChuBERT$_{\textrm{GU}}$} &  0.302 $\pm$  0.017 &  0.652 $\pm$  0.006 &  0.730 $\pm$  0.018 \valscore{&  0.436 $\pm$  0.012} &  \epochNumberStartingFromOne{36.600} $\pm$  2.221\\ 
\textbf{SChuBERT$_{\textrm{SPECTER2.0}}$} &  0.319 $\pm$  0.016 &  0.675 $\pm$  0.007 &  0.711 $\pm$  0.017 \valscore{&  0.460 $\pm$  0.003} &  \epochNumberStartingFromOne{13.400} $\pm$  6.186\\ 
\rowcolor[HTML]{d1d1d1}
 & & & \valscore{&} &  \\
 \rowcolor[HTML]{d1d1d1}
 \multirow{-2}{3.9cm}{\textbf{Multi-SChuBERT$_{\textrm{GU\_SPECTER2.0}}$}} &  
 \multirow{-2}{*}{\textbf{0.335 $\pm$  0.020}} &  \multirow{-2}{*}{\textbf{0.643 $\pm$  0.017}} &  \multirow{-2}{*}{\textbf{0.695 $\pm$  0.021}} \valscore{&  \multirow{-2}{*}{\textbf{0.484 $\pm$  0.009}}} &  \multirow{-2}{*}{\epochNumberStartingFromOne{31.300} $\pm$  11.898}\\ 
\bottomrule
}}
\end{table*}

\begin{table*}
\centering
\caption{
SPECTER2.0 results citation task, with filtered ACL training, validation and test data. 
} 
\label{table:different-embedding-results-all-data-filtered}

\scalebox{0.88}{
\mainresultstableACL{
\midrule
\rowcolor[HTML]{d1d1d1}
\textbf{Avg Training Label} & -0.001 $\pm$  0.000 &  0.861 $\pm$  0.000 &  1.046 $\pm$  0.000 & -0.000 $\pm$  0.000 &   -- \\ 
\textbf{SChuBERT} & 0.305 $\pm$  0.008 &  0.682 $\pm$  0.004 &  0.726 $\pm$  0.008 \valscore{&  0.266 $\pm$  0.004} &  \epochNumberStartingFromOne{13.700} $\pm$  4.270\\ 
\textbf{MultiSChuBERT$_{\textrm{GU}}$} &  0.332 $\pm$  0.024 &  0.647 $\pm$  0.018 &  0.698 $\pm$  0.025 \valscore{&  0.296 $\pm$  0.017} &  \epochNumberStartingFromOne{29.400} $\pm$  8.733\\ 
\rowcolor[HTML]{d1d1d1}
\textbf{SChuBERT$_{\textrm{SPECTER2.0}}$}  & 0.333 $\pm$  0.011 &  0.672 $\pm$  0.005 &  0.697 $\pm$  0.011 \valscore{&  0.325 $\pm$  0.005} &  \epochNumberStartingFromOne{17.100} $\pm$  5.238\\ 
& & & \valscore{&} &  \\
\multirow{-2}{3.9cm}{\textbf{Multi-SChuBERT$_{\textrm{GU\_SPECTER2.0}}$}} & 
\multirow{-2}{*}{\textbf{0.351 $\pm$  0.026}} &  \multirow{-2}{*}{\textbf{0.646 $\pm$  0.026}} &  \multirow{-2}{*}{\textbf{0.679 $\pm$  0.027}} \valscore{&  \multirow{-2}{*}{\textbf{0.336 $\pm$  0.009}}} &  \multirow{-2}{*}{\epochNumberStartingFromOne{22.200} $\pm$  13.831}\\ 
\bottomrule
}}
\end{table*}

\subsection{Domain-specialized embedding results} 
\label{subsection:domain_specialized_embeddings_results}

In this subsection we look at the effect of using domain-specialized embedding models inside the SChuBERT (sub-) models, in place of the BERT$_{\textrm{BASE}}$ embedding. In all the experiments discussed in this subsection, the $(u, v, |u - v|,u * v)$ concatenation method was used for the MultiSCHuBERT models, as it was found to be the best method on the ACL data with BERT$_{\textrm{BASE}}$ embeddings. 
Table \ref{table:different-embedding-results-original-data} 
shows the results for different embedding models on the original ACL data.  

The potential of using scientific-document-tailored text embeddings in our tasks is clearly hinted by these results.
However, we expected a significant overlap between the ACL data and the data used for the training and validation of these  embedding models for the science domain. In cases where such an overlap exists, we arrive in the situation of ``label leakage'', where we no longer know if improvements are due to improved models, or just the unfair advantage of having seen parts of the test data during the training of the text embedding model. We want to avoid this situation, and we can do so at least for the SPECTER2.0 embedding model, as discussed next.

\subsubsection{Eliminating label leakage}

For the SPECTER2.0 embedding model, the data used for training and validation is explicitly mentioned.\footnote{Following the information available from the HuggingFace website, \url{https://huggingface.co/allenai/SPECTER2.0} we found a series of URLs with the used data under \url{https://ai2-s2-research-public.s3.us-west-2.amazonaws.com/scirepeval} used for the training and validation of the model. Using ``wget'' these json data files can be readily downloaded.} Downloading this data in the form of a set of json files, we obtained a lists of used paper titles from it which we then used to check for overlap with the ACL data. 
As it turns out, the overlap is substantial, see Table \ref{table:ACL_datasets_label_statistics}. For example, for the ACL test set, 622 of 1548 or 40.2\% of the examples were used in the SPECTER2.0 training and validation data, with similar overlap percentages for the validation and training sets.  
Note that a similar overlap, while not easy to check, should be expected for the scibert and SPECTER embeddings,  but not BERT$_{\textrm{BASE}}$ embeddings
(since BERT$_{\textrm{BASE}}$ was trained only on BookCorpus and Wikipedia). Of these results, only the original BERT$_{\textrm{BASE}}$ results are therefore ``fair''.

We address the overlap between the ACL data and the data used for training and validation of the SPECTER2.0 embedding model. We do this by filtering the ACL data used for testing (and optionally training and validation) of our models, to eliminate the overlap. As it turns out, such filtering can be done only reasonably for the SPECTER2.0 model, because it is the only one that lists the data that has been used during its training and validation.\footnote{It should be possible to do controlled experiments for some of the other embedding models as well, which do mention the used data albeit imprecisely, such as SCINCL and SPECTER. However, this would require retraining these models in order to exactly  know the used data and avoid overlap with the test data, which would need a large amount of computation. We therefore leave this for future work.}
Fortunately, as it turns out, amongst the domain specialized embedding models SPECTER2.0 performs best, so obtaining ``fair'' scores for this model should give a reasonable idea of the kind of improvements that could be expected for the other models as well.

\subsubsection{Label Statistics original and Filtered ACL Data}
Table \ref{table:ACL_datasets_label_statistics} shows the sizes and average labels of the original ACL dataset and the variant that is created by filtering out all examples that overlap with the training or validation data used for training the SPECTER2.0 text embedding model. 
There are two main observations: 
\begin{enumerate}
 \item Within the original and filtered dataset the average and standard deviation of the label values for the different subsets is comparable, but between the datasets they differ substantially. 
\item The average and standard deviation of the label values are considerably lower within the filtered dataset.

\end{enumerate}

The lower standard deviation of label values in the filtered dataset means that this data is ``easier'' in real terms. 
Furthermore, the $R^{2}$ metric normalalizes the scores using the variance of the test set, meaning that
relatively lower $R^{2}$ scores can be expected for this second dataset given its label statistics.

\subsubsection{Results}

We now discuss the results of two slightly different approaches do eliminate the problem of label leakage.

\newchunk{Results with Filtered Test Data}
In order to restore fairness to the results, strictly speaking we are only required to exclude fro the test set examples used for the training and validation of SPECTER2.0 embedding model. The training and validation sets can remain unchanged, and we don't need to retrain the models, but can just use the original models on the new filtered test set. The results of this simple approach are shown in  Table \ref{table:different-embedding-results-original-data-filtered-testset}. Note that the systems using the SPECTER2.0 embeddings outperform those using the BERT$_{\textrm{BASE}}$ baseline embedding model. However, we also see a substantial drop in performance when comparing the validation scores to to those on the test set. The lower $R^{2}$ scores are partially due to the lower label variance in the filtered test set, see Table \ref{table:ACL_datasets_label_statistics}.
To further understand the causes of the lower results, consider the scores for the average training label prediction baseline (Avg Training Label) in  Table \ref{table:different-embedding-results-original-data-filtered-testset}. These scores show that just following the label statistics of the training data, and always predicting its average label, gives  a negative $R^{2}$ score of -0.130. This is because the label distribution of the training set  is no  longer aligned with the new test set. Therefore, obtaining high performance on the new test set given the same training data is harder, and overall lower performance is to be expected.

\newchunk{Results with Filtered  Training, Validation and Test Data}
Noting the markedly different label distribution between the original ACL training and validation sets and the filtered ACL testset, we hypothesized it  to be possible to improve performance by assuring the training, validation and test sets to have more comparable label distributions. To do so, we filtered not only the test set, but also the training and validation sets for overlap with the SPECTER2.0 training and validation data. The result is training, validation and test sets with similar label statistics, see Table \ref{table:ACL_datasets_label_statistics}. This is arguably a more coherent setup and evaluation. While it leads to smaller training and validation sets as a result of the filtering, these have statistical properties that are representative for those of the test set, which was earlier not the case.

Table \ref{table:different-embedding-results-all-data-filtered} shows the results for the new setup. Note that in this setting, we only use about 60\% of the original ACL training data, which itself can be expected to reduce performance. However, as can be seen from the results, not the $R^{2}$ and $MAE$ scores are actually higher in this new setup, which we attribute to the training and validation data that are now coherent with the test set.

\newchunk{Conclusion}
We see a substantial improvement from using the domain-specialized SPECTER2.0 embeddings over using the general-purpose BERT$_{\textrm{BASE}}$ embeddings.
Specifically, we see improvements for the
SChuBERT$_{\textrm{SPECTER2.0}}$ model compared to the 
SChuBERT (BERT$_{\textrm{BASE}}$) baseline model; as well as for the 
Multi-SChuBERT$_{\textrm{GU\_SPECTER2.0}}$ compared to the 
Multi-SChuBERT$_{\textrm{GU}}$  (BERT$_{\textrm{BASE}}$) baseline model. Importantly, these improvements are retained after eliminating the problem of label leakage.
Furthermore, if we filter each of the training and validation sets using the criterion of no overlap with SPECTER2.0 training or validation data, we obtain training, validation and test sets that are coherent in terms of label distributions. Using those new datasets we see the same kinds of improvements, while also obtaining comparatively better performance in proportion to the now substantially smaller training data.

%% file: Conclusion.tex
\acsun{SDQP}
\section{Conclusion}
In this work, we proposed a new powerful multimodal model for \ac{SDQP} by combining the SChuBERT textual model with a visual INCEPTION model. The resulting model, MultiSChuBERT obtains state-of-the-art results on the \ac{SDQP} number of citations prediction and accept/reject prediction tasks.
Our research provides several new contributions to the state-of-the-art in scholarly document quality prediction.
First, we found the visual sub-model to have a much larger capacity to fit the data in comparison to the textual model, due to substantially more trainable parameters. Given this imbalance, we show the importance of using gradual unfreezing on the visual sub-model during training, in order to get a balanced fusion of the two models and hence improved performance. Specifically, MultiSChuBERT$_{GU}$, the  MultiSChuBERT model using gradual unfreezing is shown to give the best results.
Second, we showed the importance of using the right concatenation method to combine the textual and visual embeddings in the MultiSChuBERT model.
Third, we looked at the use of domain-specialized embedding models to replace the BERT$_{\textrm{BASE}}$ embedding in the SChuBERT model/sub-model. We showed these domain-specialized embedding models, in particular SPECTER2.0, to provide substantial additional improvements on top of MultiSChuBERT$_{GU}$ using the default BERT$_{\textrm{BASE}}$ embeddings.

%% file: ChunkLengthAndOverlapAppendix.tex
\begin{table*}[t]
\centering
\scalebox{0.87}{
\begin{tabular}{lllllll}
\toprule
\textbf{Dataset} & \textbf{Majority}   & \textbf{BiLSTM} & \textbf{HAN$_\text{ST}$} &  \begin{tabular}[c]{@{}l@{}}\textbf{SChuBERT} \\ \textbf{(5 chunks)}\end{tabular} & \begin{tabular}[c]{@{}l@{}}\textbf{SChuBERT} \\ \textbf{(6 chunks)}\end{tabular} & \textbf{SChuBERT}             \\ 
\midrule
\rowcolor[HTML]{d1d1d1} 

AI      & 92.2\%        & 91.5 $\pm$ 1.03\%  & 89.6 $\pm$ 1.02\%  & 92.7 $\pm$ 1.32\% & \textbf{93.4 $\pm$ 1.18\%} &  92.4 $\pm$ 0.84\%  \\

CL      & 68.9\%      & 76.2 $\pm$ 1.30\%   & \textbf{81.8 $\pm$ 1.91\%} & 78.6 $\pm$ 3.66\% & 79.2 $\pm$ 4.01\% & 80.8 $\pm$ 2.60\% \\ 

\rowcolor[HTML]{d1d1d1} 

LG      & 68.0\%     & \textbf{81.1 $\pm$ 0.83\%}  & 78.7 $\pm$ 0.69\%  & 77.5 $\pm$ 2.78\% & 78.7 $\pm$ 1.28\% & 80.2  $\pm$ 1.39\%\\ 

\bottomrule
\end{tabular}
}
\caption{Results for the textual models for the accept/reject prediction task.}
\label{table:textual_ar}
\end{table*}

\section{Additional Experiments Chunk Properties}
\label{Additional-Experiments-Appendix}

This and the following appendices report on the results of additional experiments that were performed earlier but fall outside the scope of the main work of this paper. We note that these experiments were done using the old Tensorflow implementation of the models, using slightly different preprocessing to generate the BERT text embedding input. Absolute accuracy scores are therefore not directly comparable to those of the SChuBERT models reported in the rest of the paper. The results reported in appendices are based on averages over three runs.

\subsection{Sequence Length and Overlap}
\label{Sequence_Length_and_Overlap_Appendix}
To evaluate the effect of the chunk sequence length and overlap on performance, embeddings are generated for sequence lengths of 128, 256 and 512. Furthermore, for the 512 length embedding, two variants are generated: 1) with an overlap of 50 tokens, 2) without any overlap. Since embedding generation takes a long time, experiments are only performed on the LG dataset, which is the largest and most balanced of the PeerRead subsets. 

\subsubsection{Results}
\label{seq_len_results}
The results for the chunk sequence length and overlap experiments are shown in Table \ref{table:seqlen}. The best performing strategy is a sequence length of 512 tokens without overlap. Performance gradually decreases when the sequence length is shortened.

\begin{table}[H]
\centering
\scalebox{0.9}{
\begin{tabular}{lll}
\toprule
\textbf{Seqlen} & \textbf{Overlap} & \textbf{Accuracy}               \\ 
\midrule
\rowcolor[HTML]{d1d1d1} 

512    & Yes     & 79.48 $\pm$ 0.92\%   \\

512    & No      & \textbf{80.16 $\pm$ 1.39\%}  \\ 

\rowcolor[HTML]{d1d1d1} 

256    & No      & 78.68 $\pm$ 1.63\%  \\ 

128    & No      & 76.56 $\pm$ 1.25\%  \\

\bottomrule
\end{tabular}
}
\caption{Results for the LG dataset with chunk sequence lengths of 512, 256 and 128 and and a comparison of using an overlap of 50 tokens between chunks.}
\label{table:seqlen}
\end{table}

\subsection{PeerRead Chunk Limit Experiments}
\label{peerread_chunk_limit_experiments}

In these experiments, the SChuBERT model was evaluated by training with a portion of the chunks. Note that these experiments are similar to those of \cite{van_Dongen_2020} for number of citation prediction, but here conducted for the PeerRead dataset. The BiLSTM and HAN models are both capped at 20k characters for computational reasons, while the SChuBERT model can train with any number of characters. To ensure a fair comparison, results for 5 and 6 chunks are also included since 20k characters falls somewhere between these numbers of chunks. 

\subsubsection{Results}
\label{textual_accept_reject_results}
The results for the textual accept/reject prediction task are shown in Table \ref{table:textual_ar}, where the SChuBERT model is compared against other textual baselines on the PeerRead datasets. The SChuBERT model outperforms the other baselines on the AI domain, which is extremely unbalanced. The best performing version is SChuBERT with 6 chunks. On the other domains, it does not outperform the best performing textual baseline, but the results are stabler. For example, the BiLSTM model, which is the best performing model on the CL domain, has a performance of 0.9\% higher then SChuBERT on but 4.6\% lower on the LG domain. The HANst model, which is the best performing model on the CL domain, has a performance of 1\% higher than SChuBERT but 1.5\% lower on the LG domain. Furthermore, both best performing models perform below the majority class prediction baseline on AI, showing that SChuBERT is robust to different datasets and imbalanced classes. 

%% file: MultitaskLearningAppendix.tex
\section{Multitask Learning Experiments}
\label{Multi-task-Learning-Appendix}

\begin{figure*}[t]
    \centering
    \scalebox{0.7}{
    \includegraphics{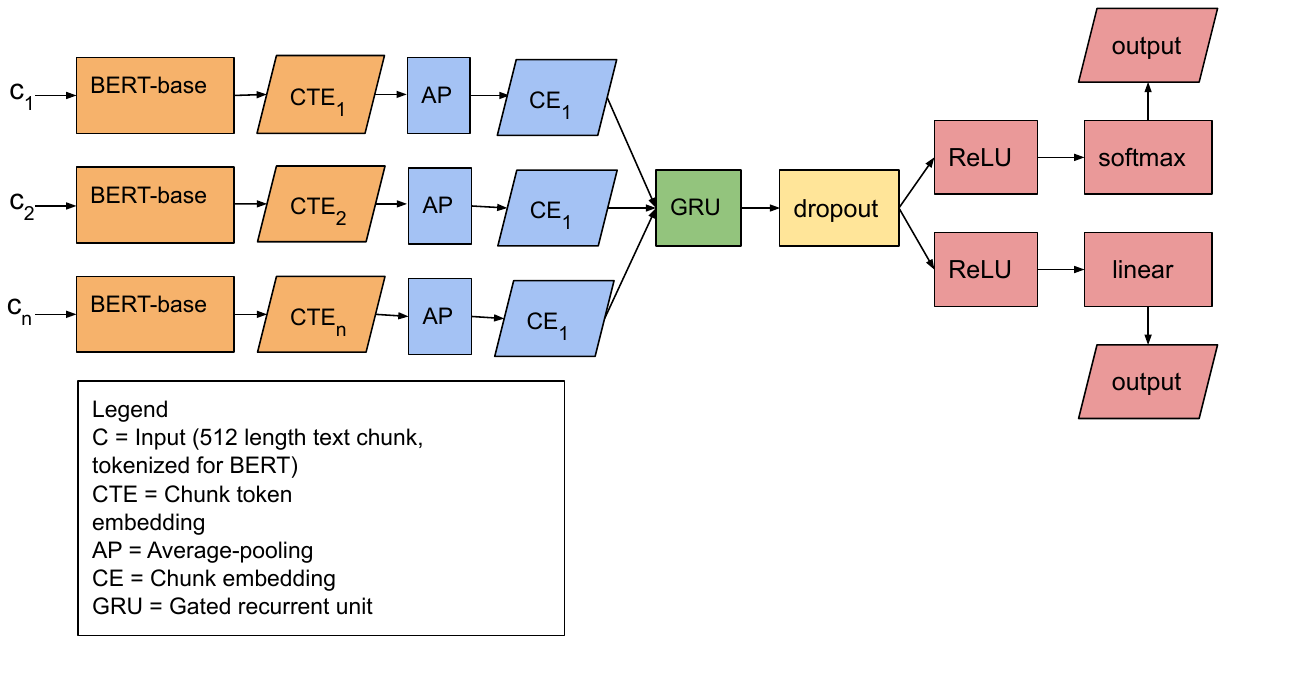}
    }
    \caption{The SChuBERT$_\text{MTL}$ model proposed in this work.}
    \label{fig:schubert-mtl-model}
\end{figure*}

The multi-task learning model used in this work to assess whether the accept/reject and citation prediction tasks can be used together in one model to improve performance is shown in figure \ref{fig:schubert-mtl-model}. It is very similar to the SChuBERT model, except for the last layers and is hence named SChuBERT$_\text{MTL}$. Each task has one task-specific fully-connected layer with a ReLU activation, followed by a fully-connected layer which leads to a softmax or linear layer. For each of the outputs, the loss is computed. The losses are then summed into a combined loss which is used to optimize the model. 

\begin{table*}[t]
\begin{minipage}{0.61\textwidth}

\centering
\scalebox{0.8}{
\begin{tabular}{lll}
\toprule
\begin{tabular}[c]{@{}l@{}}\textbf{\#Documents}\\ \textbf{(train + validation + test)}\end{tabular} & \textbf{Content}               & \textbf{Labels}                   \\ 
\midrule
31668 (28473 + 1588 + 1607)                                                    & Parsed full-text      & Accept/reject, Citations \\
\bottomrule 
\end{tabular}
}

\scalebox{0.8}{
\begin{tabular}{lll}
\toprule
 \begin{tabular}[c]{@{}l@{}}\textbf{Train} \\ \textbf{Accept : Reject}\end{tabular} & \begin{tabular}[c]{@{}l@{}}\textbf{Validation} \\ \textbf{Accept : Reject}\end{tabular}               & \begin{tabular}[c]{@{}l@{}}\textbf{Test} \\ \textbf{Accept : Reject}\end{tabular}                   \\ 
\midrule

 51.2\% : 48.8\%                                                    & 53.8\% : 46.2\%      & 52.4\% : 47.6\% \\
\bottomrule
\end{tabular}
}
\captionof{table}{Statistics of the AAPR dataset. }
\label{aaprdataset}
\end{minipage}
\begin{minipage}{0.38\textwidth}
\scalebox{0.77}{
\begin{tabular}{ll}
\toprule
\rowcolor[HTML]{FFFFFF} 
\midrule
\rowcolor[HTML]{D1D1D1} 
Vocabulary size & 30000                                                             \\
\rowcolor[HTML]{FFFFFF} 
Optimizer & Adam                                                              \\
\rowcolor[HTML]{D1D1D1} 
Learning rate (AR, CIT)   & 0.005                                                             \\
\rowcolor[HTML]{FFFFFF} 
Epochs      & 60                                                                \\
\rowcolor[HTML]{D1D1D1} 
Loss function (AR, CIT) & CE, MAE                                                           \\
\rowcolor[HTML]{FFFFFF} 
Weight initialization  & Xavier normal                                                     \\
\rowcolor[HTML]{D1D1D1} 
Dropout rate    & 0.5                                                          \\
\rowcolor[HTML]{FFFFFF} 
GRU hidden size   & 256                                                               \\
\rowcolor[HTML]{D1D1D1} 
MTL dense hidden size   & 256                                                               \\
 \rowcolor[HTML]{FFFFFF} 
\begin{tabular}[c]{@{}l@{}}Train batch size\\ \end{tabular}  & \begin{tabular}[c]{@{}l@{}} 26\end{tabular} \\
\rowcolor[HTML]{D1D1D1} 
\begin{tabular}[c]{@{}l@{}}Val batch size\\ \end{tabular} & \begin{tabular}[c]{@{}l@{}} 26\end{tabular} \\
\rowcolor[HTML]{FFFFFF}
\begin{tabular}[c]{@{}l@{}}Test batch size\\ \end{tabular} & \begin{tabular}[c]{@{}l@{}} 22\end{tabular} \\
\rowcolor[HTML]{D1D1D1} 
Word embedding size  & 768  \\
\bottomrule
\end{tabular}
}
\captionof{table}{Hyperparameters of the multi-task learning models. 'AR, CIT' refers to the accept/reject prediction and citation prediction tasks.}
\label{table:hyperparameters-mtl}
\end{minipage}
\end{table*}

\begin{table*}[t]
\centering
\scalebox{0.87}{
\begin{tabular}{llllll}
\toprule
    & \textbf{AR} & \textbf{CIT} & \textbf{MTL} & \begin{tabular}[c]{@{}l@{}}\textbf{MTL} \\ \textbf{(weighted for AR)}\end{tabular} & \begin{tabular}[c]{@{}l@{}}\textbf{MTL} \\ \textbf{(weighted for CIT)}\end{tabular}             \\ 
\midrule
\rowcolor[HTML]{d1d1d1} 

Accuracy & \textbf{63.6 $\pm$ 0.54\%} & N/A   & 63.3 $\pm$ 0.39\% & 63.4 $\pm$ 0.70\%                                                           & 63.1  $\pm$ 0.80\%    \\

R2       & N/A  & \textbf{0.164 $\pm$ 0.004} & 0.159 $\pm$ 0.013  & 0.135 $\pm$ 0.016                                                           & 0.148  $\pm$ 0.004     \\ 

\rowcolor[HTML]{d1d1d1} 

MSE      & N/A  & \textbf{1.429 $\pm$ 0.007} & 1.442 $\pm$ 0.002 & 1.480 $\pm$ 0.027                                                           & 1.457  $\pm$ 0.008       \\ 

MAE      & N/A  & \textbf{0.949 $\pm$ 0.002} & 0.953 $\pm$ 0.007 & 0.968 $\pm$ 0.010                                                          & 0.957 $\pm$ 0.002                                                           \\ 

\bottomrule
\end{tabular}
}
\caption{Results for the multi-task learning learning experiments.}
\label{table:mtl_results}
\end{table*}

\subsection{AAPR Dataset}
\label{AAPR Dataset}
The Automatic Academic Paper Rating (AAPR) dataset is a dataset proposed in \cite{AAPR}. Unfortunately, the dataset provided by the authors does not match the once described in their paper and we were unable to contact the authors and ask for clarification on this issue. Therefore, the results in their paper cannot be used as a baseline. However, the provided AAPR dataset still contains full-text information for 31668 papers with accept/reject decisions for all of them, based on whether it was accepted at a top-tier conference. This dataset only includes parsed versions of all documents and not the PDF's. Thus, this dataset is only suitable for textual models. Citation labels are also manually added to the dataset, making it the only large dataset which includes both accept/reject as well as citation labels. 
The statistics of the AAPR dataset are shown in Table \ref{aaprdataset}.

 \subsection{Method}
 \label{Multi-task Learning experiments}
The AAPR dataset is used to evaluate whether MTL can improve performance on the accept/reject prediction and citation prediction tasks by learning both tasks in one model simultaneously. Intuitively, these two tasks are related. As shown in \cite{structure_tags}, there is a large correlation between the two labels. Since the AAPR dataset includes both accept/reject as well as citation labels (after manually adding them), it is a suitable dataset to assess whether a multi-task learning model can improve performance on both tasks. As a baseline, the same model is trained on both tasks separately, since the goal is to improve performance on both tasks by training them simultaneously. The results are then compared to the MTL variant. In addition to this, an experiment is also performed to assess whether one task can improve performance on the other. For this, a model is trained where the contribution of the loss of the secondary task is halved, and the best model is selected based on the lowest loss for the primary task (instead of the combined loss).

\subsubsection{Hyperparameters}
The hyperparameters used for the multi-task learning models are shown in Table \ref{table:hyperparameters-mtl}.

\subsection{Results}
\label{mtl_results}

The results for the Multi-task Learning experiments are shown in table \ref{table:mtl_results}. The best performing model for accept/reject prediction is the AR model while the best performing model for citation prediction is the CIT model. The MTL models are not able to improve upon the single task models, but the results are very close. Weighting the losses and selecting the best model for one task decreases performance.

%% file: SChuBERTAppendix.tex
\FloatBarrier

\section{SChuBERT Results}
\label{appendix:SChuBERT-Results}
\FloatBarrier

\begin{table*}[t]
\begin{minipage}{0.58\textwidth}

\centering
\scalebox{0.90}{
\begin{tabular}{llll}
\toprule
\rowcolor[HTML]{FFFFFF} 
&  \begin{tabular}[c]{@{}l@{}}\textbf{SChuBERT} \\ \textbf{(5 chunks)}\end{tabular} & 
\begin{tabular}[c]{@{}l@{}}\textbf{SChuBERT} \\ \textbf{(6 chunks)}\end{tabular} & \textbf{SChuBERT}             \\ 
\rowcolor[HTML]{d1d1d1} 
$R^2$ score  & 0.369 $\pm$ 0.009 & 0.380 $\pm$ 0.004  & \textbf{0.398 $\pm$  0.006}   \\ 
MSE  & 1.032 $\pm$ 0.015 & 1.013 $\pm$ 0.006 &   \textbf{0.985 $\pm$   0.010}  \\ 
\rowcolor[HTML]{d1d1d1} 
MAE & 0.805 $\pm$ 0.005 & 0.798 $\pm$ 0.005 & \textbf{0.789 $\pm$ 0.005} \\ 
\end{tabular}

}
\captionof{table}{Results for the textual models for the citation prediction task.}
\label{table:textual_cit}
\end{minipage}
\begin{minipage}{0.40\textwidth}
\centering
\scalebox{0.90}{
\begin{tabular}{lll}
\toprule
  &   \begin{tabular}[c]{@{}l@{}}\textbf{SChuBERT} \\ \textbf{50\% data} \end{tabular} &   \begin{tabular}[c]{@{}l@{}}\textbf{SChuBERT} \\ \textbf{10\% data} \end{tabular}  \\
\midrule
\rowcolor[HTML]{d1d1d1} $R^2$ score & 0.327  $\pm$  0.007 & 0.205 $\pm$  0.026  \\ 
MSE & 1.058 $\pm$  0.011 & 1.473 $\pm$  0.048  \\ 
\rowcolor[HTML]{d1d1d1} MAE & 0.809 $\pm$ 0.005  & 0.923 $\pm$  0.027  \\

\bottomrule
\end{tabular}
}
\captionof{table}{Results for SChuBERT for the citation prediction task on a subset of the data and with full input.}
\label{table:textual_cit_lessdata}
\end{minipage}
\end{table*}

\begin{table*}[t]
\centering
\begin{tabular}{llll}
\toprule

  & \textbf{BiLSTM} & \textbf{HAN} & \textbf{SChuBERT} \\
\midrule
\rowcolor[HTML]{d1d1d1} 

$R^2$ score & 0.158   $\pm$ 0.006 &  0.248 $\pm$ 0.014  & \textbf{0.249 $\pm$  0.002}   \\

MSE & 1.377 $\pm$ 0.010  & 1.230 $\pm$ 0.023  &  \textbf{1.230 $\pm$   0.004}  \\ 

\rowcolor[HTML]{d1d1d1} 

MAE & 0.933 $\pm$ 0.002  & 0.885 $\pm$ 0.008  & \textbf{0.884 $\pm$ 0.002}  \\ 

\bottomrule
\end{tabular}
\caption{Results for the textual models for the citation prediction task using only abstract text.}
\label{table:textual_cit_abstractonly}
\end{table*}

This appendix repeats some of the additional results reported by 
\citet{van_Dongen_2020}, which provide details on the performance of earlier textual models on the number of citations prediction task.
These additional results are shown in Table \ref{table:textual_cit}, Table \ref{table:textual_cit_lessdata} and Table \ref{table:textual_cit_abstractonly}.